%% file: aaai24.tex
\documentclass[letterpaper]{article} % DO NOT CHANGE THIS
\usepackage{aaai24}  % DO NOT CHANGE THIS
\usepackage{times}  % DO NOT CHANGE THIS
\usepackage{helvet}  % DO NOT CHANGE THIS
\usepackage{courier}  % DO NOT CHANGE THIS
\usepackage[hyphens]{url}  % DO NOT CHANGE THIS
\usepackage{graphicx} % DO NOT CHANGE THIS
\usepackage{natbib}  % DO NOT CHANGE THIS AND DO NOT ADD ANY OPTIONS TO IT
\usepackage{caption} % DO NOT CHANGE THIS AND DO NOT ADD ANY OPTIONS TO IT
\usepackage{algorithm}
\usepackage{algorithmic}

\usepackage{newfloat}
\usepackage{listings}
\DeclareCaptionStyle{ruled}{labelfont=normalfont,labelsep=colon,strut=off} % DO NOT CHANGE THIS
\floatstyle{ruled}
\newfloat{listing}{tb}{lst}{}
\floatname{listing}{Listing}
\newtheorem{theorem}{Theorem}
\usepackage{amssymb}
\usepackage{amsmath}
\usepackage{mathtools}
\usepackage{latexsym}
\usepackage{lineno}
\usepackage{nccmath}
\usepackage{adjustbox}
\usepackage{booktabs}
\usepackage{multirow}
\usepackage{cleveref}
\usepackage{lipsum}
\usepackage{verbatim}
\newcommand{\Tau}{\mathcal{T}}

\usepackage{color}
\definecolor{amber}{rgb}{1.0, 0.75, 0.0}
\definecolor{green}{rgb}{0.0, 0.5, 0.0}

\newcommand {\yulunliu}[1]{{\color{red}\textbf{Yu-Lun: }#1}\normalfont}

\newcommand{\etal}{\textit{et al.}}
\title{Improving Robustness for Joint Optimization of Camera Poses \\and Decomposed Low-Rank Tensorial Radiance Fields}
\author {
    Bo-Yu Cheng,
    Wei-Chen Chiu,
    Yu-Lun Liu
}
\affiliations {
    National Yang Ming Chiao Tung University\\
    tomas1999.ee06@nycu.edu.tw, walon@cs.nctu.edu.tw, yulunliu@cs.nycu.edu.tw
}

\begin{document}

\maketitle

\begin{abstract}

In this paper, we propose an algorithm that allows joint refinement of camera pose and scene geometry represented by decomposed low-rank tensor, using only 2D images as supervision. 
% First, we derive and analyze the frequency property of the 1D signal to show that naive joint pose optimization on voxel-based NeRFs can easily lead to sub-optimal solutions.
First, we conduct a pilot study based on a 1D signal and relate our findings to 3D scenarios, where the naive joint pose optimization on voxel-based NeRFs can easily lead to sub-optimal solutions.
% Based on frequency property analysis on the 1D signal alignment, we propose a separable component-wise convolution mechanism that enables camera pose alignment in 2D and 3D radiance fields. 
Moreover, based on the analysis of the frequency spectrum, we propose to apply convolutional Gaussian filters on 2D and 3D radiance fields for a coarse-to-fine training schedule that enables joint camera pose optimization.
% a separable component-wise convolution mechanism that enables  in 2D and 3D radiance fields. 
Leveraging the decomposition property in decomposed low-rank tensor, our method achieves an equivalent effect to brute-force 3D convolution with only incurring little computational overhead. 
To further improve the robustness and stability of joint optimization, we also propose techniques of smoothed 2D supervision, randomly scaled kernel parameters, and edge-guided loss mask. Extensive quantitative and qualitative evaluations demonstrate that our proposed framework achieves superior performance in novel view synthesis as well as rapid convergence for optimization.
The source code is available at https://github.com/Nemo1999/Joint-TensoRF.
% without known camera poses.

% % This sentence is same as robust-ngp abstract (need to rewrite)
% Experiments on novel view-synthesis datasets validate that our learning framework achieves state-of-the-art performance and rapid convergence of neural rendering, even when initial camera poses are unknown.
\end{abstract}

\input{CameraReady/LaTeX/intro}
\input{CameraReady/LaTeX/related}

\input{CameraReady/LaTeX/method}
\input{CameraReady/LaTeX/exp}

\input{CameraReady/LaTeX/conclusion}

\section*{Acknowledgments}
This work is supported by National Science and Technology Council (NSTC) 111-2628-E-A49-018-MY4, 112-2221-E-A49-087-MY3, 112-2222-E-A49-004-MY2, and Higher Education Sprout Project of the National Yang Ming Chiao Tung University, as well as the Ministry of Education (MoE), Taiwan. In particular, Yu-Lun Liu acknowledges the Yushan Young Fellow Program by the MoE in Taiwan.

% \clearpage

\bibliography{CameraReady/LaTeX/aaai24}

\end{document}

% \newpage
% \maketitle
\onecolumn
\begin{center}
\textbf{\LARGE{Improving Robustness for Joint Optimization of Camera Poses \\
and Decomposed Low-Rank Tensorial Radiance Fields
\\
\textit{Supplementary Material}
}
\vspace{1em}
\\
\Large{Anonymous submission}
}
\end{center}
\vspace{2em}
\normalsize

\section{Overview}
This supplementary material presents additional results to complement the main manuscript. First, we provide details on the spectrum analysis of 1D signal alignment in Sec.~\ref{sec:analysis_1d}. 
% Then, we describe the experimental setup, including the datasets, compared methods, and evaluation metrics in~\secref{setup}. Next, we show the complete visual comparisons of the camera pose estimation results on the MPI Sintel Dataset in~\secref{visual_sintel}. 
Second, we describe the connection between 1D and 3D Gaussian filtering, which is the main motivation behind our proposed method, in Sec.~\ref{sec:gaussian_1d_3d}.
Next, we derive the equivalence of separable component-wise convolution and na\"ive 3D convolution in Sec.~\ref{method_3D}.
Finally, we provide the complete training process and implementation details in Sec.~\ref{sec:training_process} and Sec.~\ref{exp:implementation}, respectively.
Alongside this document, we provide additional video results and compare our results with state-of-the-art methods.
% an interactive HTML interface to compare our video results with state-of-the-art methods.

\section{Complete Version of Spectrum Analysis of 1D Signal Alignment }
\label{sec:analysis_1d}

In Eq. 5 of Sec. 3.2, we formulated the 1D signal alignment problem as: 
% Notice the similar structure of the formulation of $\mathcal{L}_\text{1d}$ Eq.~\ref{eq:1d_loss} and Eq.~\ref{eq:joint3D}.
\begin{equation}
    \label{eq:sup_1d_loss}
    \begin{split}
         {\mathcal{L}_\text{1d}(g,q_1, q_2)}  &  { = \sum_{i\in [1,2]} \int \lVert \mathcal{W}_\text{1d}(g, q_i)(x) - f_i(x) \rVert^2 dx }\\
        & {=  \sum_{i\in [1,2]} \int \lVert g(x) - f_{GT}(x- p_i + q_i) \rVert^2 dx},\\
    % \text{where }   \mathcal{W} \text{ is the } & \text{translation operator on signal defined as} \\
    %      {\mathcal{W}_\text{1d}(g, q_i)(x) }&  {= g(x - q_1)}        
        %g^*, q_1^*, q_2^* & = \operatorname*{argmin}_{g, q_1, q_2}  \mathcal{L}_\text{1d}(g, q_1, q_2)
    \end{split}
\end{equation}
where    $\mathcal{W}$ is the translation operator on signal defined as ${\mathcal{W}_\text{1d}(g, q_i)(x) }  {= g(x - q_1)}$, $f_{GT}$  is the ground truth 1D signal, $p_1, p_2$ are predetermined translation values, and $g$ is the signal we are trying to optimize, and $q_1, q_2$ are the reconstructed translation values. We optimize $\mathcal{L}_\text{1d}$ with gradient descent and attempt to reach one of the global minima, where $q_1 - q_2 = p_1 - p_2$ and $g$ = $\mathcal{W}_\text{1d}(f_{GT}, p_1-q_1)$

\subsection{Theorem 1: Simplifying Joint Optimization to Pure Alignment on Shifted GT Signals}
To simplify the complex dynamic interaction of joint optimization in Eq. \ref{eq:sup_1d_loss}, here we assume that $g$ rapidly converges to temporary optima (with respect to the current translation) before the translation parameters $q_1, q_2$ are further refined. Note that this assumption is reasonable because voxel-based architectures (which we focus on) easily overfit to current supervision.  Due to the property that the minima of squared error are achieved at average value, this allows us to replace $g$ by the average of two translated ground truth signals. (Note that the partial optimization of $g$ is convex and is guaranteed to converge to $g^*_{q_1, q_2}$).  

\begin{equation}
    \label{eq:sup_g_star}
    \begin{split}
        g^*_{q_1, q_2} & = \operatorname*{argmin}_{g}  \mathcal{L}_\text{1d}(g, q_1, q_2) \\
                       & = \operatorname*{argmin}_{g}  \sum_{i\in [1,2]} \int \lVert g(x) - f_{GT}(x- p_i + q_i) \rVert^2 dx \\
                       & = \frac{f_{GT}(x- p_1 + q_1)+f_{GT}(x- p_2 + q_2)}{2}.
    \end{split}
\end{equation}

Substituting Equation \ref{eq:sup_g_star}, the loss $\mathcal{L}_{1d}$ with respect to $q_1$ and $q_2$ under this assumption can be simplified as follows.

\begin{equation}
    \label{eq:sup_align_1D}
    \begin{aligned}
        \mathcal{L}_\text{1d}(& q_1, q_2)  = \mathcal{L}_\text{1d}(g^*_{q_1, q_2},q_1, q_2) \\
                         & = \sum_{i\in [1,2]} \int \lVert g^*(x-q_i) - f_i(x) \rVert^2 dx \\
                         & =   \sum_{i\in [1,2]} \int \lVert g^*(x)_{q_1, q_2} - f_{GT}(x- p_i + q_i) \rVert^2 dx  \\
                         & =    {\sum_{i\in [1,2]} \int \lVert \frac{f_{GT}(x-p_1+q_1) - f_{GT}(x- p_2 + q_2)}{2} \rVert^2 dx} \\
                         & =   \int \lVert  {f_{GT}(x-p_1+q_1) - f_{GT}(x- p_2 + q_2)} \rVert^2 dx \\
                         & =   \int \lVert  {f_{GT}(x) - f_{GT}(x+u) } \rVert^2 dx \\
                         & =   \int \lVert  {f_{GT}(x) - \mathcal{W}(f_{GT},-u)} \rVert dx,
    \end{aligned}
\end{equation}
where $u = (p_1-p_2) - (q_1-q_2)$ is the translation value of the pure alignment problem between two shifted ground truth signals $f_{GT}$ and $\mathcal{W}(f_{GT},-u)$. 

\subsection{Theorem 2: Spectral Property of Gradient in 1D Signal Alignment}

\begin{equation}
    \label{eq:sup_1d_fourier}
    \begin{aligned}
        \text{Let } u       & =  {(p_1-p_2) - (q_1-q_2)}                                        && \text{(a). replace variable}\\
        \mathcal{L}_{1d} & =  \int \lVert  {f_{GT}(x) - f_{GT}(x+ u) } \rVert^2 dx && \\
               & =  \int \lVert \  {\mathfrak{F}[\ {f_{GT}(x) - f_{GT}(x+ u) }\ ]} \ \rVert^2 dk && \text{(b). Parseval's theorem}\\ 
               & =  \int \lVert \  {\mathfrak{F}[\ {f_{GT}\ ]} -  e^{iku}\mathfrak{F}[\ f_{GT}\ ]} \ \rVert^2 dk && \text{(c). shift property} \\
               & =  \int \lVert \   {\mathfrak{F}[\ {f_{GT}\ ]} }\ \rVert^2  \cdot \lVert \  {1-e^{iku}}\  \rVert^2 dk \\
        \frac{d}{d u}\mathcal{L}_{1d} & =  \frac{d}{d u} \int \lVert \  {\mathfrak{F}[\ {f_{GT}\ ]}} \ \rVert^2  \cdot \lVert \ 
         {1-e^{iku}} \ \rVert^2 dk\\
                            & =   \int \lVert \  {\mathfrak{F}[\ {f_{GT}\ ]}}\  \rVert^2  \cdot \frac{d}{d u} \lVert \  {1-e^{iku}} \ \rVert^2 dk  && \text{(d). swap operators by Leibniz integral rule }\\ 
                            & =   \int \lVert \  {\mathfrak{F}[\ {f_{GT}\ ]}} \ \rVert^2  \cdot H(u,k) \ dk ,
              %               \\
              % \text{where }& H(u,k)  = 4 \pi k \  sin( 2 \pi k u),
    \end{aligned}
\end{equation}
where $H(u,k)  = 4 \pi k \  sin( 2 \pi k u)$, $\mathfrak{F}[\ f_{GT} \ ]$ represents the Fourier transform of the spatial domain function $f_{GT}(x)$, and $k$ is the wavenumber in the frequency domain. 
%In step (b) above, we apply Parseval's theorem, which states that the energy in the spatial domain is the same as the energy in the frequency domain. In step (d) above, we are allowed to swap the order of derivative and integration because $f_{GT}$ satisfies the criteria of the Leibniz integral rule. 
In the final form, the function $H(u,k)$ transfers the spectrum $\mathfrak{F}[\ f_{GT}\ ]$ into the derivative $\frac{d}{du}\mathcal{L}_{1d}$. The value of $H(u,k)$ is plotted in the left part of Fig. \ref{fig:sup:1D_H_function}. We can see that the sign of the transfer function $H$ is well-behaved when the magnitude of $k$ is small. Here ``well-behaving'' means that the sign of the gradient is able to help $u$ descend to $0$. (i.e., positive when $u>0$ and negative when $u<0$). However, when the magnitude of $k$ increases, the sign of $H$ quickly begins to alternate with increasing magnitude. If the spectrum of $f_{GT}$ is too wide, the sign of $\frac{d}{du}\mathcal{L}_{1d}$ will be affected by the flipping sign of $H(u,k)$, and joint optimization will get stuck in local optima. 

\subsection{Theorem 3: Effect of Gaussian Kernel on the 1D Signal Alignment}
\begin{figure*}[th!]
    \centering
    \includegraphics[width=0.9 \textwidth]{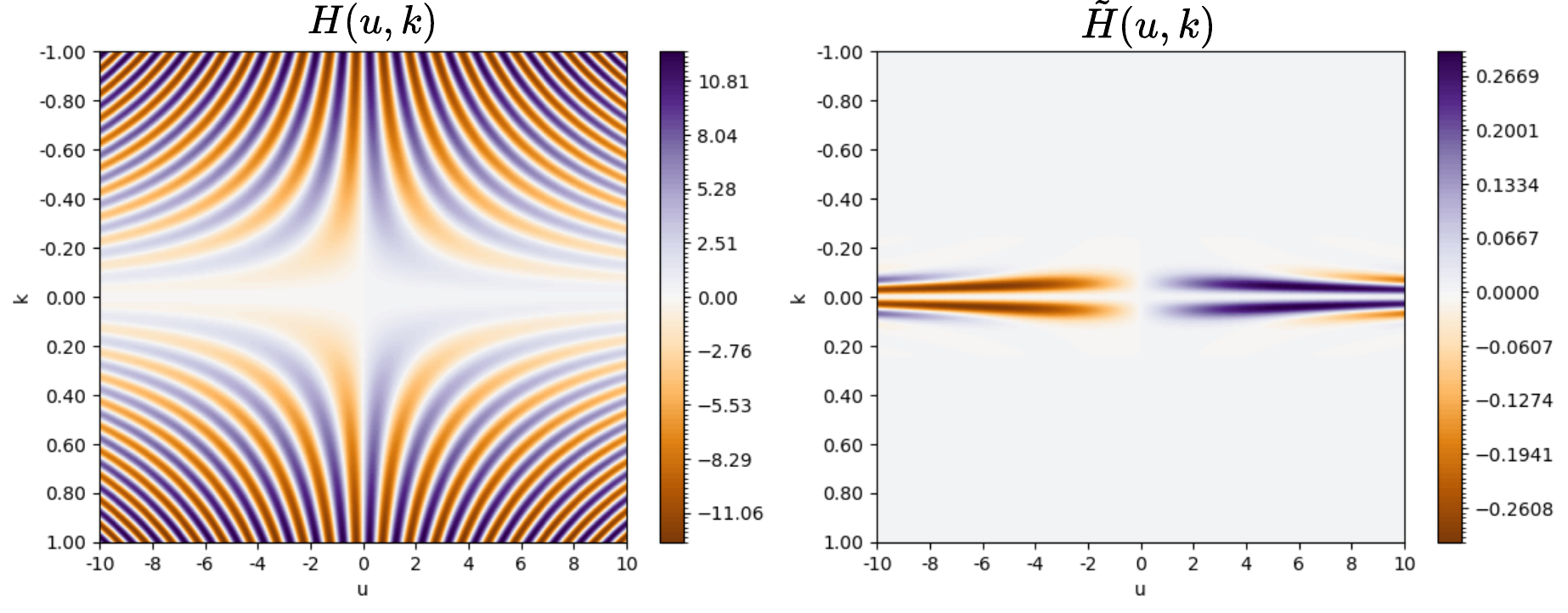}
    \caption{\textbf{Copy of Fig. 3(b)} is placed here for readability. Left: The value of $H(u,k)$ is plotted. We can see that the sign of the transfer function $H$ is well-behaved when the magnitude of $k$ is small. Here, ``well-behaving'' means that the sign of the gradient is able to help $u$ descend to $0$, i.e., positive when $u>0$ and negative when $u<0$. However,  when $k$ departs from $0$, the sign of $H$ quickly begins to alternate and the magnitude increases, which causes the gradient to be large and noisy. Hence, high-frequency signals with a spreading spectrum can easily get stuck in local optima. Right: Filtering the $f_{GT}$ with Gaussian kernel is effectively the same as modulating the function $H(u,k)$ with a Gaussian mask, resulting in a more desirable function $\Tilde{H}(u,k)$}.
    \label{fig:sup:1D_H_function}
\end{figure*}

To deal with this flipping gradient issue of $H(u,k)$ when $k$ departs from $0$, we try to shrink the bandwidth of $f_{GT}$ so that it won't reach too much into the flipping area of the transfer function $H$. Applying the 1D Gaussian filter $\mathcal{N}(x)$ on $f_{GT}$ is a natural solution. Here we show that applying a Gaussian filter to $f_{GT}$ is effectively the same as modulating the transfer function $H$.
\begin{equation}
    \label{eq:1D_filter}
    \begin{aligned}
            \text{Let } \Tilde{f}_{GT} & =\mathcal{N} \ast_\text{1d} \Tilde{f}_{GT} \text{ be the filtered signal}\\
            \Tilde{\mathcal{L}}_{1d}             & = \mathcal{L}_{1d} \text{ calculated with } \Tilde{f}_{GT}\\ 
            \frac{d}{d u}\Tilde{\mathcal{L}}_{1d} & =  \int \lVert \ \mathfrak{F}[\ \mathcal{N} \ast_\text{1d} f_{GT}\ ] \ \rVert^2  \cdot H(u,k) \ dk && \text{(a) Substitute Eq. \ref{eq:sup_1d_fourier}}\\
                                & = \int \lVert \ \mathfrak{F}[\ \mathcal{N} \ ] \cdot \mathfrak{F}[ \ f_{GT}\ ] \ \rVert^2  \cdot H(u,k) \ dk && \text{(b) convolution property}\\
                                & =  \int \lVert \ \mathfrak{F}[\ f_{GT} \ ] \rVert^2 \cdot \lVert \  \mathfrak{F}[ \ \mathcal{N} \ ] \ \rVert^2 \cdot H(u,k) \ dk &&\text{(c) } \mathfrak{F}[\ \mathcal{N}\ ] \text{is real}\\
                                & =   \int \lVert \ \mathfrak{F}[\ f_{GT} \ ] \rVert^2 \cdot \Tilde{H}(u,k) \ dk,
                                % \\
            % \text{where } &  \Tilde{H}(u,k) = \rVert \ \mathfrak{F}[\ \mathcal{N} \ ] \lVert ^2 \cdot H(u,k) 
    \end{aligned}
\end{equation}
where $\Tilde{H}(u,k) = \rVert \ \mathfrak{F}[\ \mathcal{N} \ ] \lVert ^2 \cdot H(u,k)$, and  $\ast_\text{1d}$ denotes the 1D convolution operator. Step (a) comes directly from Equation \ref{eq:sup_1d_fourier}, and step (b) applies the convolution property of the Fourier transform. In step (c), we use the identity $\mathfrak{F}[\mathcal{N}(x)] = \mathfrak{F}[e^{-ax^2}] = -\sqrt{\frac{\pi}{a}} e^{- \pi ^2 k^2 / a}$, which says that the Fourier transform of Gaussian function is another Gaussian function; hence $\mathfrak{F}[\mathcal{N}(x)]$ is real-valued and can be separated out of the 2-norm.

In Fig 3(b)(bottom) in the main paper (same as the right half of Figure~\ref{fig:sup:1D_H_function}), we plot the modulated transfer function $\Tilde{H}(u,k)$ (filter $N(x)$ is generated by Eq.11 with $\sigma = 4.0$). We can see that the misbehaved region is suppressed and the gradient descent is very likely to converge to $u=0$ as long as the initial magnitude of $u$ is less than $6.0$. (Actually, the region in which $\frac{d}{du} \Tilde{\mathcal{L}}_{1d}$ is well-behaved is \emph{quasi-convex} and is guaranteed to converge to global optima given the suitable learning rate that prevents us from getting stuck at saddle points.)

\section{Connection of Gaussian Filtering in 1D and 3D \\(Motivation for Proposed Techniques In Sec. 3.6)}
\label{sec:gaussian_1d_3d}
Here we discuss the connections between the 1D analysis in Sec.3.2 and the 3D joint optimization techniques proposed in Sec. 3.6.
\subsection{Connection to Smoothed 2D Supervision \& 3D Gaussian Filtering}
Note that in the previous section, we only considered blurring the input signals $f_{GT}$ (which in turn affects $f_1$ and $f_2$ in Eq. \ref{eq:sup_1d_loss}). If we strictly map this setting into the joint optimization in the 3D case as described in Sec. 3.2.2, we should only blur 2D input training images (\emph{Smoothed 2D Supervision} in Sec. 3.6). However, we found empirically that restricting the spectrum of both 2D training images and the 3D radiance field gives the best reconstruction quality (as shown in Tab. 4).

\subsection{Motivation for Randomly Scaled Kernel \& Edge Guided Loss}
\begin{figure*}[h!]
    \centering
    \includegraphics[width=1.0 \textwidth]{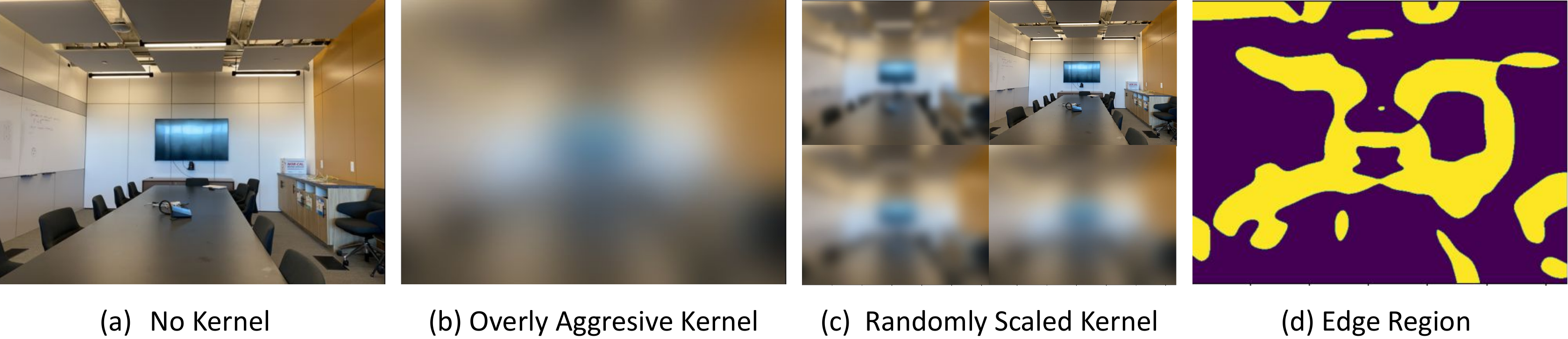}
    \caption{\textbf{Visualization of 2D Randomly Sampled Kernel and Edge Guided Loss}. (a) Input supervision without kernel. Joint optimization using unblurred images easily overfit to high-frequency noises (b) Input supervision blurred by an overly aggressive kernel. Notice that the edge information is largely destroyed by the blurring process, resulting in weak and noisy gradients, causing the poses to drift around easily. (c) Same input supervision blurred by four randomly scaled kernels. We empirically found that mixing different filtering strengths results in a more robust joint optimization. (d) Edge areas of the blurred image selected by the Sobel filter with a threshold set to 1.25x of the average value of the filtered map.}
    \label{fig:randomscale_edge_vis}
\end{figure*}

From the previous spectral analysis, one may have the impression that a larger kernel leads to stronger modulation, and hence always results in more robust pose registration. However, this is not always true, because the magnitude of $H(u,k)$ decreases linearly as $k$ approaches $0$. Notice that in Fig. \ref{fig:sup:1D_H_function}(Right) the magnitude of modulated $\Tilde{H}$ is weaker than that of $H$, which means that $\frac{d}{du}\Tilde{\mathcal{L}}_\text{1d}$ is weaker than $\frac{d}{du}\mathcal{L}_\text{1d}$ and therefore is more easily influenced by noise. In the 3D case, this \textbf{weak and noisy gradient problem} caused by overly aggressive filtering corresponds to the excessive blur effect that destroys important edge signals in the training images, causing pose alignment to fail. See Fig. \ref{fig:randomscale_edge_vis}(b) for a visualization of the image blurred by an over-strength kernel, in which the thin edge information is eliminated, causing the camera pose to randomly drift.

Real-world scenes are composed of edge structures of various scales; it is insufficient to use a single-size kernel on all these different scene structures (in which the same kernel may be overly aggressive in one scene, but overly gentle in another scene). Therefore, we introduce \emph{randomly scaled kernel} in Sec. 3.6, which randomly scales the kernel by a factor uniformly sampled from $[0,1]$. See Fig \ref{fig:randomscale_edge_vis}(c) for a visualization of the same input image filtered by a range of randomly sampled kernels. We observe that the training schedule becomes more robust when we alternate between these randomly sampled kernel scales.

Another way to mitigate the weak and noisy gradient problem is the \emph{edge guided loss} introduced in Sec. 3.6, in which we increase the learning rate (and hence amplify the gradient signal) on pixels in the edge area. See visualization in Fig. \ref{fig:randomscale_edge_vis} (d), where we color the edge area (which is detected using the Sobel filter on the filtered 2D images) in yellow. Edge-guided rendering loss helps joint optimization focus more on the edge areas of the training images, resulting in more robust pose optimization.

\section{Theorem 4: Equivalence of Separable Component-Wise Convolution and Naive 3D Convolution} 
\label{method_3D}
In Eq. 15 of Sec. 3.5, we used the following identity to simplify the computation of Naive 3D convolution into separable component-wise convolution. 

\begin{equation}
    \begin{aligned}
        \Tilde{\Tau_\sigma} & = 
                     \sum_{r=1}^{\mathbf{R}} {\mathbf{\Tilde v}^X_{\sigma, r}} \otimes {\mathbf{\Tilde M}^{Y,Z}_{\sigma,r}} + 
                                       {\mathbf{\Tilde v}^Y_{\sigma, r}} \otimes {\mathbf{\Tilde M}^{X,Z}_{\sigma,r}} + 
                                       {\mathbf{\Tilde v}^Z_{\sigma, r}} \otimes {\mathbf{\Tilde M}^{X,Y}_{\sigma,r}},
    \end{aligned}
    \label{eq:separate_goal}
\end{equation}
where $\Tilde \Tau_\sigma = (\mathcal{N}_{\text{3d}} \ast_{\text{3d}} \Tau_\sigma)$ denotes the 3D Gaussian convoluted tensor volume, ${\mathbf{\Tilde v}_{\sigma, r}} = (\mathcal{N}_{1d} \ast_{1d} \mathbf{\Tilde v}_{\sigma,r})$ denotes the 1D Gaussian convoluted vector component, and ${\mathbf{\Tilde M}_{\sigma,r}} = (\mathcal{N}_{\text{2d}} \ast_{\text{2d}} \mathbf{\Tilde M}_{\sigma, r})$ denotes the 2D Gaussian convoluted matrix component. 

To prove Eq. \ref{eq:separate_goal}, we first notice that the outer product between the lower-dimensional component $\otimes$ can be viewed as convolution $\ast$ in a higher-dimensional space. We can view the outer product as a convolution with full padding, where the kernel and the signal are embedded into the high-dimensional space before performing the convolution. For example, the outer product between two 1D vectors $\mathbf{u} \in \mathbb{R}^n$ and $\mathbf{v} \in \mathbb{R}^m$ is a $m \times n$ matrix. We can obtain the same result by performing 2D convolution on 2D discrete embedded functions ${\mathbf{u}}^X, {\mathbf{v}}^Y: \mathbb{Z}^2 \to \mathbb{R}$.
\begin{equation}
    \begin{aligned}
        {\mathbf{u}}^X[x,y] &= \begin{cases}
            \mathbf{u}[x]\mathbf{\delta}[y] & \text{ if } 0 \leq x  < n\\
            0  & \text{otherwise}
        \end{cases}\\
        {\mathbf{v}}^X[x,y] &= \begin{cases}
            \mathbf{v}[y]\mathbf{\delta}[x] & \text{ if } 0 \leq y  < m\\
            0  & \text{ otherwise}    
        \end{cases}\\
        (\mathbf{u}^X \ast_\text{2d} \mathbf{v}^Y)[x,y] &= \sum_{i=-\infty}^{\infty} \sum_{j=-\infty}^{\infty} \mathbf{u}^X[i,j]\cdot \mathbf{v}^Y[x-i, y-j] && \text{(a) expand definition}\\
            &= \sum_{i=-\infty}^{\infty} \sum_{j=-\infty}^{\infty} \mathbf{u}[i]\ \delta[j] \cdot \mathbf{v}[y-j]\ \delta[x-i] && \text{(b) only add turns where } j=0\text{ and }  x-i=0\\
            &= \mathbf{u}[x]\ \delta[0] \cdot \mathbf{v}[y-0]\ \delta[0] &&\text{(c) substitute values}\\ 
            &= \mathbf{u}[x] \cdot \mathbf{v}[y]\\
            &= (\mathbf{u}\otimes \mathbf{v})[x,y], && \text{(e) definition of }\otimes
    \end{aligned}
    \label{eq:sup_outer_cov_2d}
\end{equation}
where in (c) we assume $\mathbf{u}[x]$ and $\mathbf{v}[y]$ returns $0$ when index is out of range of the vector. Now we have proved that the result of the outer product $\otimes$ between 1D vectors is identical to the convolution $\ast_\text{2d}$ of two embedded vectors in the 2D space.  

Similar properties can be obtained for vector-matrix products: 
\begin{equation}
    \begin{aligned}
        (\mathbf{v}^{X} \ast_\text{3d} \mathbf{M}^{Y,Z})[x,y,z] = (\mathbf{v} \otimes \mathbf{M})[x,y,z], && \text{(a) proof is similar to Eq. \ref{eq:sup_outer_cov_2d}},
    \end{aligned}
    \label{eq:sup_outer_cov_3d}
\end{equation}
where $\mathbf{v}^X$ is the embedded version of the 1D vector $\mathbf{v}$ along the $X$ axis in the 3D space, and $\mathbf{M}^{Y,Z}$ is the embedded version of the 2D matrix $\mathbf{M}$ along the $Y,Z$ axes in the 3D space. 

Now with Eq. \ref{eq:sup_outer_cov_2d} and Eq. \ref{eq:sup_outer_cov_3d} we can replace $\otimes$ in 3D Gaussian definition and Eq. \ref{eq:separate_goal} by convolution $\ast_\text{3d}$, after which we apply the commutative and associative property of the convolution operator $\ast_\text{3d}$ to distribute the separable Gaussians into each tensor component.

\begin{equation}
    \begin{aligned}
        \Tilde{\Tau}_\sigma & = \mathcal{N}_\text{3d} \ast_\text{3d} \Tau_\sigma \\
                            & = (\mathcal{N}_\text{1d}^X \otimes \mathcal{N}_\text{1d}^Y   \otimes \mathcal{N}_\text{1d}^Z) \ast_\text{3d} 
                            (\sum_{r=1}^{\mathbf{R}} 
                                \mathbf{v}^X_{\sigma,r} \otimes \mathbf{M}^{Y,Z}_{\sigma, r} +
                                \mathbf{v}^Y_{\sigma,r} \otimes \mathbf{M}^{X,Z}_{\sigma, r} + 
                                \mathbf{v}^Z_{\sigma,r} \otimes \mathbf{M}^{X,Y}_{\sigma, r}
                            ) &&\text{(a) expand definition}\\
                            & =  (\mathcal{N}_\text{1d}^X \ast_\text{3d} \mathcal{N}_\text{1d}^Y   \ast_\text{3d} 
                            \mathcal{N}_\text{1d}^Z) \ast_\text{3d} 
                            (\sum_{r=1}^{\mathbf{R}} 
                                \mathbf{v}^X_{\sigma,r} \ast_\text{3d} \mathbf{M}^{Y,Z}_{\sigma, r} +
                                \mathbf{v}^Y_{\sigma,r} \ast_\text{3d} \mathbf{M}^{X,Z}_{\sigma, r} + 
                                \mathbf{v}^Z_{\sigma,r} \ast_\text{3d} \mathbf{M}^{X,Y}_{\sigma, r}
                            ) &&\text{(b) apply Eq. \ref{eq:sup_outer_cov_2d} and Eq. \ref{eq:sup_outer_cov_3d}}\\
                            & = 
                            \sum_{r=1}^{\mathbf{R}} 
                                (\mathcal{N}_\text{1d}^X \ast_\text{3d} \mathcal{N}_\text{1d}^Y   \ast_\text{3d} \mathcal{N}_\text{1d}^Z) \ast_\text{3d} (\mathbf{v}^X_{\sigma,r} \ast_\text{3d} \mathbf{M}^{Y,Z}_{\sigma, r}) + \\
                                & \quad \quad \quad  (\mathcal{N}_\text{1d}^X\ast_\text{3d} \mathcal{N}_\text{1d}^Y   \ast_\text{3d} \mathcal{N}_\text{1d}^Z) \ast_\text{3d} (\mathbf{v}^Y_{\sigma,r} \ast_\text{3d} \mathbf{M}^{X,Z}_{\sigma, r}) + \\
                                & \quad \quad \quad  (\mathcal{N}_\text{1d}^X \ast_\text{3d} \mathcal{N}_\text{1d}^Y   \ast_\text{3d} \mathcal{N}_\text{1d}^Z) \ast_\text{3d} (\mathbf{v}^Z_{\sigma,r} \ast_\text{3d} \mathbf{M}^{Z,Y}_{\sigma, r})
                                &&\text{(c) linearity of } \ast_\text{3d}\\
                            & = 
                            \sum_{r=1}^{\mathbf{R}} 
                                {(\mathcal{N}_\text{1d}^X \ast_\text{3d} \mathbf{v}^X_{\sigma,r}) \ast_\text{3d} ((\mathcal{N}_\text{1d}^Y   \ast_\text{3d} \mathcal{N}_\text{1d}^Z) \ast_\text{3d} \mathbf{M}^{Y,Z}_{\sigma, r}) +} \\
                                & \quad \quad \quad  {(\mathcal{N}_\text{1d}^Y \ast_\text{3d} \mathbf{v}^Y_{\sigma,r}) \ast_\text{3d} ((\mathcal{N}_\text{1d}^X   \ast_\text{3d} \mathcal{N}_\text{1d}^Z) \ast_\text{3d} \mathbf{M}^{X,Z}_{\sigma, r}) +} \\
                                & \quad \quad \quad  {(\mathcal{N}_\text{1d}^Z \ast_\text{3d} \mathbf{v}^Z_{\sigma,r}) \ast_\text{3d} ((\mathcal{N}_\text{1d}^X   \ast_\text{3d} \mathcal{N}_\text{1d}^Y) \ast_\text{3d} \mathbf{M}^{X,Y}_{\sigma, r})} 
                                &&\text{(c) commutivity , associativity, of } \ast_\text{3d}\\
                            & = \sum_{r=1}^{\mathbf{R}} 
                                       {\mathbf{\Tilde v}^X_{\sigma, r}} \ast_\text{3d} {\mathbf{\Tilde M}^{Y,Z}_{\sigma,r}} + 
                                       {\mathbf{\Tilde v}^Y_{\sigma, r}} \ast_\text{3d} {\mathbf{\Tilde M}^{X,Z}_{\sigma,r}} + 
                                       {\mathbf{\Tilde v}^Z_{\sigma, r}} \ast_\text{3d} {\mathbf{\Tilde M}^{X,Y}_{\sigma,r}} && \text{(d) definition rewrite}\\
                            & = \sum_{r=1}^{\mathbf{R}} 
                                       {\mathbf{\Tilde v}^X_{\sigma, r}} \otimes {\mathbf{\Tilde M}^{Y,Z}_{\sigma,r}} + 
                                       {\mathbf{\Tilde v}^Y_{\sigma, r}} \otimes {\mathbf{\Tilde M}^{X,Z}_{\sigma,r}} + 
                                       {\mathbf{\Tilde v}^Z_{\sigma, r}} \otimes {\mathbf{\Tilde M}^{X,Y}_{\sigma,r}}. && \text{(e) apply Eq. \ref{eq:sup_outer_cov_3d}}.
    \end{aligned}
    \label{eq:seperable_conv_tensorf}
\end{equation}
Now the proof is complete, and we have verified that \textbf{the 3D convoluted tensor can be expressed as the composition of individually convoluted components}. With a similar process, we can prove that the 3D feature tensor also has the property: 
\begin{equation}
    \begin{aligned}
        \Tilde{\Tau}_c & = \mathcal{N}_\text{3d} \ast_\text{3d} \Tau_c \\
                       & = \sum_{r=1}^{\mathbf{R}} 
                                       {\mathbf{\Tilde v}^X_{c, r}} \otimes {\mathbf{\Tilde M}^{Y,Z}_{c,r}} \otimes \mathbf{b}^X + 
                                       {\mathbf{\Tilde v}^Y_{c, r}} \otimes {\mathbf{\Tilde M}^{X,Z}_{c,r}} \otimes \mathbf{b}^Y + 
                                       {\mathbf{\Tilde v}^Z_{c, r}} \otimes {\mathbf{\Tilde M}^{X,Y}_{c,r}} \otimes \mathbf{b}^Z . 
    \end{aligned}
\end{equation}

\section{Complete Training Process}
\label{sec:training_process}
The training loss of our method in joint optimization of the 3D running field can be summarized as follows
\begin{equation}
    \begin{aligned}
         \Tilde{\mathcal{L}}_\text{joint}(F_\sigma, F_c, \mathbf{P}, \mathcal{N}_\sigma, \mathcal{N}_c, \mathcal{N}_I) & = \sum_{i =1}^L   \sum_{u\in \mathbf{U}} \mathrm{E}_iu \cdot 
        {\lVert \mathbf{V}(\Tilde{F}_\sigma, \Tilde{F}_c, \mathcal{W}_\text{3d}(P_i, s(\vec{0}, \vec{d_u}))) - 
        \Tilde{I}_{iu} \rVert} \\
        \Tilde{F}_\sigma (\textbf{x}) & = \Tilde{\Tau}_\sigma(\textbf{x}), \text{ where } \Tilde{\Tau}_\sigma \text{ is component-wise convoluted with } \mathcal{N}_\sigma  \\
        \Tilde{F}_c(\textbf{x}, \vec{d}) &= s(\Tilde{\Tau}_c(\textbf{x}), \vec{d}), \text{ where } \Tilde{\Tau}_c \text{ is component-wise convoluted with } \mathcal{N}_c  \\
        \Tilde{I} &= \{\mathcal{N}_I^X \ast_\text{2d} \mathcal{N}_I^Y \ast_\text{2d} I_i\}_{i=1,2,\cdots,L}, 
     \end{aligned}
     \label{eq:sup_blurred_3D_joint_loss}
\end{equation}
where $E_{iu}$ is the edge-guided rendering weight of pixel $u$ on image $i$ (In the first few iterations, $E_{iu}=1.5$ for pixels on the edge regions, after that, $E_{iu}=1.0$ for all training pixels.), $\mathcal{N}_c, \mathcal{N}_\sigma$ are 1D Gaussian kernel for convolving $\Tau_c, \Tau_\sigma$ respectively, $\mathcal{N}_I$ is the 1D Gaussian kernel that is used for smoothing the supervision images $I$.

The complete training loss can finally be expressed as 
\begin{equation}
    \begin{aligned}
        \mathcal{L}_\text{3d} = w_1 \cdot \mathcal{L}_\text{joint} + w_2 \cdot \mathcal{L}_\text{L1} + w_3 \cdot \mathcal{L}_\text{TV},
    \end{aligned}
    \label{eq:sup_total_loss}
\end{equation}
where $\mathcal{L}_\text{joint}$ is described in Eq. \ref{eq:sup_blurred_3D_joint_loss}, $\mathcal{L}_\text{L1}, \mathcal{L}_\text{TV}$ are the L1 loss and TV loss on tensor components $\mathbf{v}_{\sigma, r}, \mathbf{M}_{c, r}, \mathbf{M}_{\sigma, r} , \mathbf{v}_{c, r}$ respectively. $w_1, w_2, w_3$ are loss weights.

Now, the total training process can be described as Algorithm \ref{alg:training}, where the kernel width is first sampled from a predefined kernel schedule, then randomly scaled by a factor sampled uniformly from $[0, 1]$, then Gaussian kernels are constructed and used to calculate our proposed rendering loss $\mathcal{L}_\text{joint}$ in Eq. \ref{eq:sup_blurred_3D_joint_loss}. Finally, the total loss $\mathcal{L}_\text{3d}$ in Eq. \ref{eq:sup_total_loss}, based on which the camera poses and the radiance field are jointly updated. 

\begin{algorithm}[h!]
\begin{algorithmic}
\STATE $\Tau_\sigma, \Tau_c \leftarrow $ Initialize Voxel Grid
\STATE $\mathbf{P} \leftarrow$ Initialize Camera Poses
\FOR{$s$ = 1 \TO train\_iters}
\STATE  $i \leftarrow$ 2D\_kernel\_sched($s$)
\STATE  $\sigma$, $c$ $\leftarrow$ 3D\_kernel\_sched($s$)
\STATE  $u_\sigma \leftarrow$ randomly sample density kernel scale
\STATE  $u_I \leftarrow$ randomly sample 2D kernel scale  
\STATE  $\mathcal{N}_\sigma \leftarrow$ sample discrete gaussian kernel with variance $(\sigma \cdot u_\sigma)^2$
\STATE  $\mathcal{N}_c \leftarrow$ sample discrete gaussian kernel with variance $c^2$
\STATE  $\mathcal{N}_I \leftarrow$ sample discrete gaussian kernel with variance $(i \cdot u_I)^2$
\STATE  $\Tilde{\mathcal{L}}_\text{joint} \leftarrow \Tilde{\mathcal{L}}_\text{joint}(\Tau_\sigma, \Tau_c, \mathbf{P}, \mathcal{N}_\sigma, \mathcal{N}_c, \mathcal{N}_I)$ (with randomly selected pixels in all training views)
\STATE  $\mathcal{L}_\text{L1}$ =  $\mathcal{L}_\text{L1}(\mathbf{v}_{\sigma, r}, \mathbf{M}_{c, r}, \mathbf{M}_{\sigma, r} , \mathbf{v}_{c, r})$
\STATE  $\mathcal{L}_\text{TV}$ =  $\mathcal{L}_\text{TV}(\mathbf{v}_{\sigma, r}, \mathbf{M}_{c, r}, \mathbf{M}_{\sigma, r} , \mathbf{v}_{c, r})$
\STATE  $\mathcal{L}_\text{3d} = w_1 \cdot \mathcal{L}_\text{joint} + w_2 \cdot \mathcal{L}_\text{L1} + w_3 \cdot \mathcal{L}_\text{TV}$
\STATE back propagation
\STATE update $\Tau_\sigma, \Tau_c, \mathbf{P}$
\ENDFOR
\end{algorithmic}
\caption{Conceptual Training Process for Our Proposed 3D Joint Optimization Training}
\label{alg:training}
\end{algorithm}

\section{Implementation Details}
\label{exp:implementation}
We evaluate our proposed method against three previous works BARF \cite{lin2021barf}, GARF \cite{chng2022gaussian}, and HASH \cite{heo2023robust}. Since the implementations of GARF and HASH are unavailable, we directly use the results reported in their paper for comparison. For the planar image alignment task, we compare our result with \cite{lin2021barf} under the same settings. To reconstruct the radiance field, we follow \cite{lin2021barf, chng2022gaussian, heo2023robust} to evaluate and compare our method on \textbf{NeRF-Synthetic} dataset and \textbf{LLFF} dataset.

\textbf{Planar Image Alignment} is performed under settings identical to those of BaRF \cite{lin2021barf}, where a sample image is chosen from ImageNet \cite{deng2009imagenet} , from which $L_\text{2d} = 5$ patches are cropped with different homography transform parameters $P_0, P_1, \cdots, P_5 \in \mathbb{R}^8$. 
We jointly reconstruct the homography parameters and the 2D image represented with the decomposed 2D tensor. The warp parameters are parameterized in $sl(3)$ and initialized to $\vec{0}$ before training.

\textbf{NeRF-Synthetic} dataset is proposed by \cite{yen2020inerf}, and consists of 8 object-centric scenes. For each scene, the training data consist of 100 images along with the camera extrinsic and intrinsic. Following \cite{lin2021barf}, we simulate camera noise with Gaussian noise $N(0, 0.15I)$ on $\mathfrak{se}(3)$ pose embedding. The noises are composed of the ground truth extrinsic parameter, and our joint optimization process aims to cancel the noise and restore accurate camera poses. We follow BaRF \cite{lin2021barf} to resize training and testing images to $400 \times 400$.

\textbf{LLFF} dataset is proposed by \cite{mildenhall2019local_llff}, and consists of 8 forward-facing scenes captured with a handheld camera. The ground truth camera poses in the datasets are estimated by COLMAP \cite{schoenberger2016mvs_COLMAP, schoenberger2016sfm_COLMAP}. Following BaRF \cite{lin2021barf}, we optimize camera poses from scratch with identity initialization, and images are resized to $480 \times 640$ before being used.

\subsection{Implementation Details for Planar Image Alignment} We use $500\times 500$ 2D decomposed tensor with $100$ components. The 2D tenor and homography warp parameters are jointly optimized with \emph{Adam} optimizer \cite{kingma2014adam} with learning rates $0.001$ and $0.01$ respectively. The total training iteration is set to $15000$, Gaussian kernel schedule shrinks exponentially with a pise-wise linear curve that starts with $128$ and reaches $0$ at $6000$ iteration. 
\begin{comment}
    In the schedule, the kernel width shrinks by $0.5$ x every $1500$ training iteration, until it is finally set to $0$ after $6000$ iterations.

    We found empirically that the optimization is robust across different kernel schedule designs ( linear, exponential, all works) as long as the initial kernel width is large enough, and the kernel width shrinks monotonically. 
\end{comment}

\subsection{Implementation Details for NeRF (3D) Synthetic Object}
We follow the implementation of \cite{chen2022tensorf} and sample 2048 rays per iteration across all training images with sample density equal to 2X current tensor grid resolution. We decrease the hidden width of the decoder \emph{ MLP} to $32$ and postpone the input of the viewing direction to the last layer. The pose parameters, the tensor volume, and the MLP decoder are jointly optimized with \emph{Adam} optimizer with learning rates $0.001$, $0.01$, and $0.0005$, respectively, and the learning rates are exponentially degraded.  \emph{Smooth 2D supervision} in Section 3.6 is used in synthetic scenes. The total training iteration is set to $40000$. The 2D and 3D Gaussian kernel schedule shrinks exponentially with a pice-wise linear curve, which starts with $0.3$ (in 3D coordinate) and $0.025$(in 2D coordinate)  and degrades to effectively $0$ at $10000$ iteration. The 3D tensor grid is scaled from $64^3$ to $300^3$ with $5$ up-sample steps, and kernel parameters are adjusted to adapt to the current tensor grid resolution. We postpone the alpha mask update in \cite{chen2022tensorf} to allow blurry scenes in the early stage of training. The total training time is $80$ minutes on a single RTX3090 GPU. 

\subsection{Implementation Details for NeRF (3D) Real World Scenes} 
Most settings in Real World Scenes are the same as the Synthetic Scenes. We describe the difference here, where we follow \cite{yen2020inerf} and \cite{chen2022tensorf} to use the Normalized Device Coordinate (NDC) in joint optimization of the forward-facing scenes. We found that the absolute coordinates (before Procruste analysis to align with GT) of reconstructed poses are often more densely gathered than GT poses, causing the absolute size of the reconstructed scene to be smaller and clipped by the near plane in NDC coordinate; we solve this issue by moving the sampling near the plane of NDC from 1.0 to -1. Most other settings are the same as the synthetic setting. The total training iteration is set to $50000$. The 3D tensor grid is scaled from $128^3$ to $800^3$ with $5$ up-sample steps. The Gaussian kernel schedule is the same as the synthetic NeRF setting, except that we start with $0.4$ in 3D coordinates for the 3D kernel and $0.07$ in 2D coordinates for the 2D kernel. To accelerate the training process in increase the stability, we increase the number of rays per iteration to $20480$ before $6000$ iterations and restore to $4096$ rays per iteration afterward. The learning rate of the pose parameters warms up for 500 iterations, and poses are reset to identity at 2500 iterations. \emph{Randomly scaled Kernel Parameters} and \emph{Edge-Guided Loss} described in Section 3.6 are used in real-world scenes. Note that to prevent input scale instability of the \emph{MLP} decoder, we use randomly scaled Gaussian kernel only for density volume, the Gaussian kernel for the color volume is not randomized. The total training time is about 3 hours on a single RTX3090 GPU.

\subsection{Evaluation Criteria}\label{exp:evalution}
Following previous works \cite{lin2021barf, chng2022gaussian, heo2023robust}, we measure our results in two aspects: pose error for registration and view-synthesis quality for the scene representation. Since the reconstructed poses and scenes are variable to the ground truth up to a similarity transform, we perform the Procrustes analysis to align the training pose parameter and the GT poses before calculating the rotation error and translation error. To report the view-synthesis quality independent of tiny registration noise, before calculating PSNR, SSIM, and LPIPS on the trained scene w.r.t testing image, we perform test-time optimization on each testing pose to prevent tiny pose error from contaminating the view-synthesis quality. In the 2D case, warp error (distance between homography parameter and the ground truth warp) and PSNR are reported.

\bibliography{CameraReady/LaTeX/aaai24}

%% file: CameraReady/LaTeX/intro.tex
\section{Introduction}

In recent years, neural rendering has become a widely-used method for high-quality novel view synthesis. NeRF as a pioneer work \cite{mildenhall2020nerf} represents a 3D radiance field as an implicit continuous function built upon multilayer perceptrons (MLPs) which is trained with differentiable volume rendering. While achieving excellent synthesis quality, NeRF suffers from training/inference inefficiency due to dense evaluation of the computationally expensive MLPs.
% during training and testing. 

To this end, voxel-based methods built upon the explicit scene representation of 3D voxel grid \cite{SunSC22_Voxel, yu2022plenoxels_voxel, liu2020neural_Voxel} are proposed to achieve faster training and provide better rendering quality than the original MLP-based NeRF, hence becoming the more preferred choices for downstream applications. 

Nevertheless, maintaining a dense 3D voxel grid is in turn memory intensive, thus still restricting wider applications of voxel-based methods. Fortunately, TensoRF \cite{chen2022tensorf} proposes to tackle such memory-intensive issue of the voxel grid via replacing the dense 3D grid with {\emph{decomposed low-rank tensor}.
TensoRF achieves a high data compression ratio and low computational cost at the same time while also achieving state-of-the-art performance. 
Providing a win-win situation on memory usage and computational efficiency, the decomposed low-rank tensor architecture has been widely adopted in many recent works \cite{xu2023avatarmav, fridovich-keilKPlanesExplicitRadiance2023, goelStyleTRFStylizingTensorial2022, hanMultiscaleTensorDecomposition2023, shaoTensor4DEfficientNeural2023, tangCompressiblecomposableNeRFRankresidual2022a, Meuleman_2023_CVPR}.

% Figure Teaser  
\begin{figure}[t]
\centering
\includegraphics[width=1.0\columnwidth]{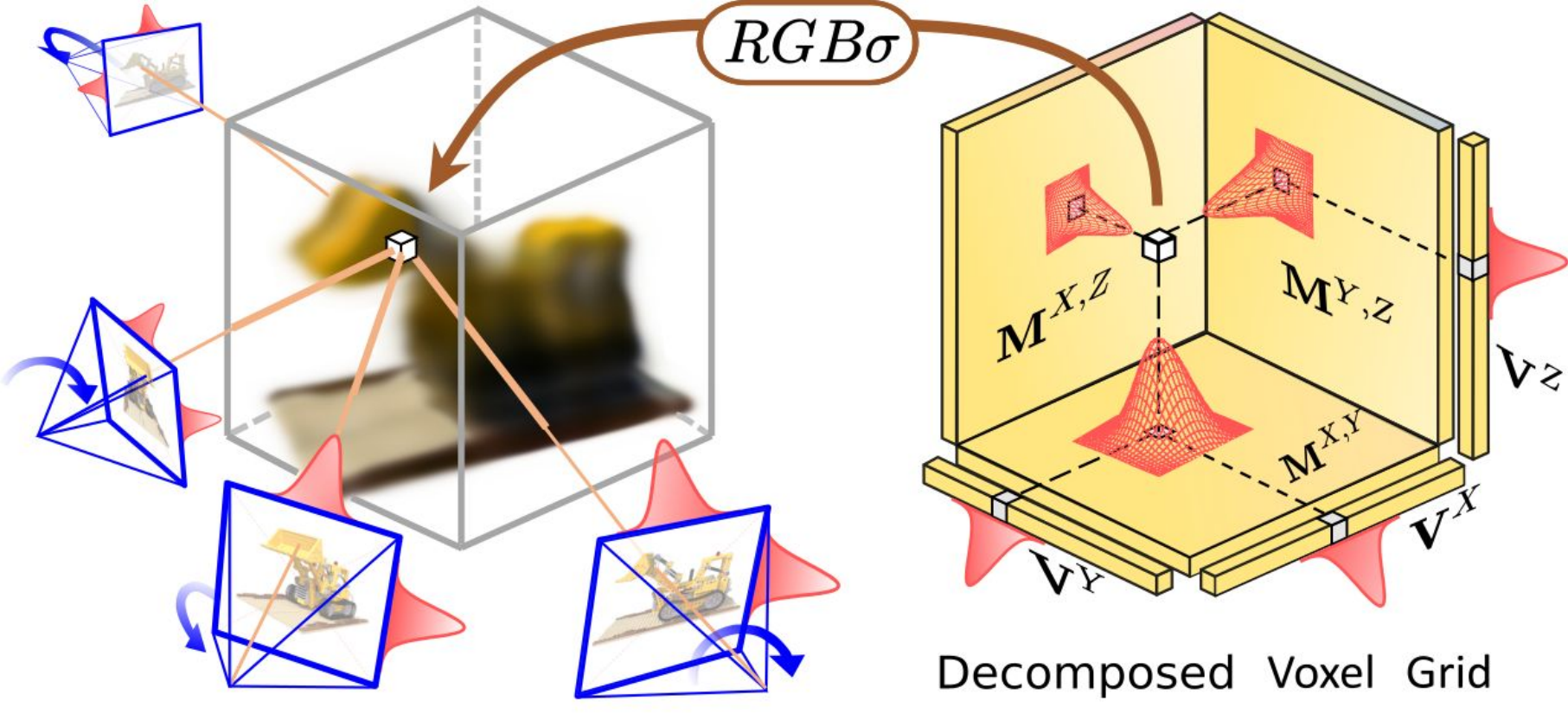} % Reduce the figure size so that it is slightly narrower than the column. Don't use precise values for figure width.This setup will avoid overfull boxes.
% \vspace{-2em}
\caption{\textbf{Robust joint pose refinement on decomposed tensor.} Our method enables joint optimization of \textbf{\color{blue}{camera poses}} and \textbf{\color{amber}{decomposed voxel representation}} by applying efficient \emph{separable component-wise convolution} of \textbf{\color{red}{Gaussian filters}} on 3D tensor volume and 2D supervision images.} 
\label{fig:teaser}
% \vspace{-1.5em}
\end{figure}

On the other hand, the effectiveness of NeRF (and most of the aforementioned works) hinges on precise camera poses of input images, which are often calculated using Structure-from-Motion (SfM) algorithms like COLMAP~\cite{schoenberger2016sfm_COLMAP}. While some works~\cite{wang2021nerfmm, lin2021barf, chng2022gaussian} aim to bypass the slow and occasionally inaccurate COLMAP process by optimizing camera pose and scene representation jointly on the original MLP-based NeRF, their success is often tied to the spectral bias~\cite{inr_dictionaries2022} of the MLP architecture which ensures the smoothness of 3D radiance field early in training. Voxel-based methods, however, lack such properties and can overemphasize sharp edges, making naive joint optimization problematic as getting trapped in local optima (Fig.~\ref{fig:naive} (a)).

% This paper proposes simple yet effective techniques for achieving joint refinement of camera pose and 3D scene on decomposed low-rank tensor (Figure ~\ref{fig:teaser}). First, we demonstrate that limiting the frequency spectrum is the key factor for pose alignment by performing spectrum analysis on the 1D signal alignment problem. Second, we propose a way to efficiently limit the frequency spectrum of decomposed low-rank tensors. While spectrum control is easy to achieve on MLP-based NeRF due to the intrinsic smoothness bias of MLP \cite{yuce2022structured}, it is tricky to restrict the spectrum of data explicitly stored in a dense 3D grid because naive filtering requires computationally expensive 3-dimensional convolution, destroying the speed advantage of voxel-based methods. We proposed a computationally efficient way to perform 3D filtering by \emph{component-wise separable convolution},  achieving spectral control on decomposed low-rank tensor.
% % ~\yulunliu{We need to be very specific in describing our proposed method in the intro section. For example, We propose a computationally effective way by introducing separable 3D Gaussian filters to blablabla.}~\boyu{fixed}
% % 
% Finally, to address the instability of the joint optimization process, we proposed \emph{smoothed 2D supervision}, \emph{randomly scaled kernel paramter}, and \emph{edge-guided loss mask} to prevent joint optimization from getting stuck in local minima. The ablation study shows that these mechanisms are essential to the success of robust pose refinement on decomposed low-rank tensors.
In this work, we present simple yet effective methods for refining the camera pose and the 3D scene using decomposed low-rank tensors (cf. Fig.~\ref{fig:teaser}). We identify that controlling the frequency spectrum is vital for pose alignment, while directly realizing such control in a dense 3D grid could be nontrivial/challenging as well as computationally demanding.
% While MLP-based NeRF naturally limits its spectrum due to the inherent smoothness bias \cite{yuce2022structured}, achieving this in a dense 3D grid is challenging and can be computationally demanding. 
To this end, we introduce an efficient 3D filtering method using \emph{component-wise separable convolution} for enabling the spectral control, which is more efficient than the traditionally well-known trick of \textit{separable convolution kernel} as we additionally utilize the separability of the input signal. To further ensure stability in the optimization process, we propose several techniques, including \emph{smoothed 2D supervision}, \emph{randomly scaled kernel paramter}, and the \emph{edge-guided loss mask}. These techniques are experimentally proven crucial for successful pose refinement in our ablation studies. In results, our proposed method requires only 50k training iterations, where all the previous methods typically needs 200k iterations (e.g. the overall training time is reduced to 25\%, compared to previous MLP-based methods). The main reason behind this advantage is not only based on property of voxel-based architecture, but also relies on our carefully designed efficient spectral filtering algorithm that requires only single reusable voxel grid (please refer to Sec. \ref{exp:time}).
Moreover, our method performs favorably against state-of-the-art methods on novel view synthesis. Our contributions are three-fold:

% In terms of novel view synthesis, we achieve state-of-the-art quality, surpassing existing methods. 
%
% Our contributions are three-fold:
% ~\yulunliu{The contribution section is pretty good!} ~\boyu{thanks for being nice}
\begin{itemize}
    % \item We propose an efficient learning strategy that jointly optimizes decomposed low-rank tensor and camera pose by leveraging separable component-wise convolution for effective and accurate control of the coarse-to-fine training schedule. 
    % \item To increase the robustness of the joint learning process, we further propose smoothed 2d supervision,  randomly scaled kernel parameter, and edge-guided loss mask to prevent the joint optimization from getting stuck in local optima.  
    % \item The training time is reduced by 25\% compared to previous MLP-based methods. The number of training iterations is only 50k compared to 200k of previous methods. Quantitative analysis demonstrates state-of-the-art performance on novel view synthesis with unknown camera pose. 
    % \item We propose a learning strategy optimizing tensor and camera pose with component-wise convolution, ensuring a coarse-to-fine training schedule. 
    % \item We derive and analyze why naive pose optimization with voxel-based radiance field reconstruction leads to sub-optimal solutions and propose a learning strategy with component-wise convolution, ensuring a coarse-to-fine training schedule. 
    \item With 1D pilot study, we provide insights into the impact of spectral property of 3D scene on the convergence of joint optimization beyond the coarse-to-fine heuristic discussed in prior research, and propose a learning strategy built upon specially designed efficient component-wise convolution algorithm. 
    \item To enhance the robustness of our joint optimization, we introduce techniques of smoothed 2D supervision, scaled kernel parameters, and the edge-guided loss mask. 
    \item Training time drops by 25\% versus existing MLP-based methods, with requiring only 50k iterations against 200k of previous methods. Results show state-of-the-art performance in novel view synthesis with unknown pose.
\end{itemize}

%% file: CameraReady/LaTeX/related.tex
\section{Related Work}
% \topic{Accelerating Neural Rendering}
% \subsection{Accelerating Neural Rendering}
% Neural Rendering, first proposed by NeRF \cite{mildenhall2020nerf}, is a state-of-the-art novel-view synthesis method. The original version of NeRF represents a 3D radiance field as a implicit continuous function with multilayer perceptron (\emph{MLP}), and trained with differentiable volume rendering.
\noindent\textbf{Accelerating Neural Rendering.}
As the seminal work of neural rendering, NeRF adopts MLPs to construct the implicit representation of the 3D scene, providing high-quality view synthesis but having a time-consuming training process due to the computational demands of MLPs. For addressing such issue, different variants of NeRF are proposed to use custom spatial data structures where the scene information is distributed only locally thus aiding faster training and rendering, in which those spatial data structures include \emph{point cloud} \cite{xu2022point, Hu_2023_CVPR_TriVolPoint}, \emph{space partitioning tree} \cite{wang2022fourier_tree, Yu2021PlenOctreesFR}, 
\emph{triangular mesh} \cite{chen2022mobilenerf_mesh, kulhanek2023tetranerf_mesh},
and \emph{voxel gird} \cite{SunSC22_Voxel, yu2022plenoxels_voxel,  liu2020neural_Voxel, hedman2021snerg}. Among these variants, the voxel grid has become more popular due to its easy implementation and quality reconstruction. However, as scene dimensions grow, the memory usage of the voxel grid becomes inefficient. To address this, \cite{mueller2022instant} recommends compressing the grid via hash encoding, while \cite{chen2022tensorf, kplanes_2023} suggest adopting tensor decomposition for 3D feature compression, in which our method is mainly based on \cite{chen2022tensorf} but can be adaptable to other tensor decomposition-based voxel structures like K-Planes~\cite{kplanes_2023}.

\noindent\textbf{Joint Pose Estimation on MLP-based NeRFs.}
% \cite{yen2020inerf} and 
\cite{wang2021nerfmm} is one of the first NeRF-based attempts to tackle the joint problem of estimating camera poses and learning 3D scene representation by directly adjusting camera pose using gradient propagation on neural radiance fields. 
% the task of jointly optimizing NeRF, 
% which directly adjusts camera pose using gradient propagation on Neural Radiance Fields.
The robustness of such joint optimization is further enhanced by \cite{lin2021barf, chng2022gaussian}, where they propose various methods to smooth the pose gradient derived from the underlying MLP. \cite{chen2023local} further increases the noise tolerance by a specially designed local-global joint alignment approach.
Our method also tackles joint problems but is specifically designed for the voxel-based NeRF built upon the decomposed low-rank tensor architecture. 
% This makes our method much faster than previous MLP-based methods while achieving state-of-the-art reconstruction quality.

% \topic{Pose Estimation on Decomposed Low-rank Tensors}
\noindent\textbf{Pose Estimation on Decomposed Low-rank Tensors.} 
There do exist works that optimize camera pose on decomposed low-rank tensor \cite{liu2023robust,Meuleman_2023_CVPR} but require rich additional geometry clues (e.g., depth map and optical flow). To our best knowledge, we are the first attempt to jointly optimize the camera pose and the \emph{decomposed low-rank tensor} using only 2D image supervision. 
% Recent works that optimize camera pose on decomposed low-rank tensor  \cite{liu2023robust} \cite{Meuleman_2023_CVPR} require rich additional geometry clues (e.g., depth map and optical flow). 
% To our knowledge, we are the first to jointly optimize the camera pose and the \emph{ decomposed low-rank tensor} using only 2D image supervision. 
% We believe that our method may be combined with existing methods to improve the local bundle adjustment quality. 

% Figure Naive Method Compare 
\begin{figure}[t!]
\centering
\includegraphics[width=\columnwidth]{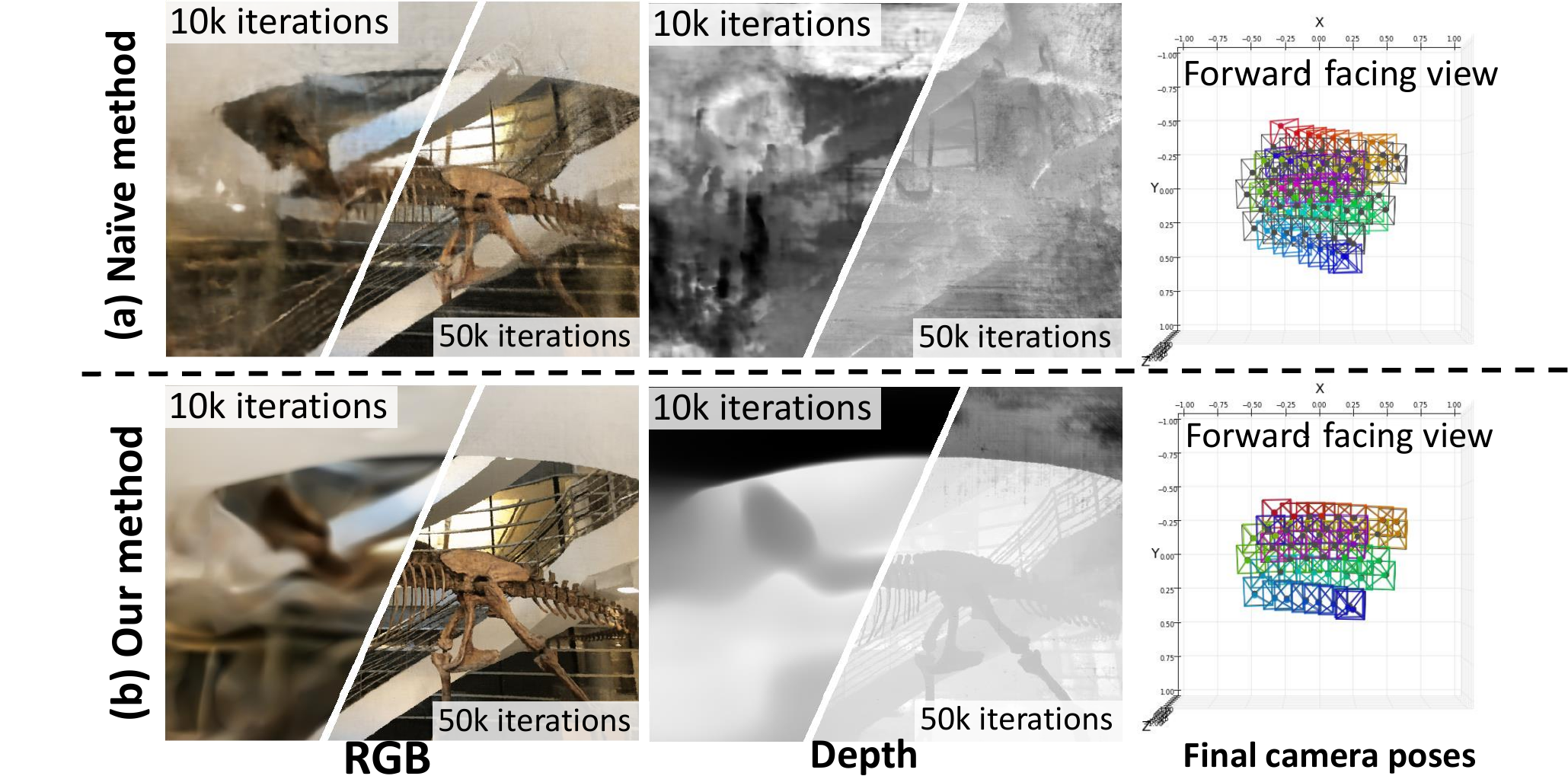}
% \vspace{-2.0em}
\caption{\textbf{Comparison of naive joint pose optimization and our proposed method on voxel-based NeRFs.} (a) Naively applying joint optimization on voxel-based NeRFs leads to dramatic failure as premature high-frequency signals in the voxel volume would curse the camera poses to stuck in local minima. (b) We propose a computationally effective manner to directly control the spectrum of the radiance field by performing \emph{separable component-wise convolution} of Gaussian filters on the decomposed tensor. The proposed training scheme allows the joint optimization to converge successfully to a better solution.}
\label{fig:naive}
% \vspace{-1.3em}
\end{figure}

% \topic{Pose Estimation on Multi-Resolution Hash Encoding}
\noindent\textbf{Pose Estimation on Multi-Resolution Hash Encoding.}
Aside from decomposed low-rank tensor, \emph{multi-resolution hash encoding} is another compressed voxel-based architecture proposed by \cite{mueller2022instant}. Along with such a choice of architecture, recently \cite{heo2023robust} has proposed to address the joint optimization of camera pose and multi-resolution hash encoding.
% with a new interpolation scheme and learning rate schedule.
% Using the parallelized multi-resolution hash table and custom \textit{nvidia cuda}\cite{nickolls2008scalable_cuda} implementation, \cite{mueller2022instant} achieves very fast training of NeRF in hundreds of seconds.
% Recently, \cite{heo2023robust} has achieved joint optimization of camera pose and multi-resolution hash encoding with a new interpolation scheme and learning rate schedule. 
They suggest a new interpolation scheme that provides smooth gradients hence preventing gradient fluctuation in the hash volume, along with a curriculum learning scheme that controls the learning rate of the hash table at each resolution level. Although achieving impressive results on joint optimization, the effectiveness of their method is limited to multi-resolution hash encoding and is not applicable to \emph{decomposed low-rank tensor},
 While our proposed separable component-wise 3D convolution (and randomly scaled kernel) is specifically designed for \emph{decomposed low-rank tensor} and not directly applicable to \emph{multi-resolution hash encoding}, in which these two representations have their respective pros and cons.

%% file: CameraReady/LaTeX/method.tex
\section{Our Proposed Method}

% In Section \ref{method_BaRF}, we formulate the joint optimization for optimizing camera poses and 3D scenes.
% In Section \ref{method_1D}, we reduce the joint optimization problem to a simpler form (i.e., 1D signal alignment) and explain how controlling the signal spectrum with Gaussian kernels can help the joint optimization.
% In Section \ref{method_2D}, we describe and test the effect of 2D Gaussian convolution on the Planar Image Alignment task. 
% Section \ref{method_TensoRF} summarizes the design of the decomposed low-rank tensor proposed by \cite{chen2022tensorf}.
% In Section \ref{method_3D}, we introduce \emph{separable component-wise convolution} on decomposed low-rank tensor, enabling us to effectively control the spectrum of the 3D radiance field.
% In Section \ref{method_Robust}, we proposed several mechanisms that increase the robustness of joint optimization. 

\subsection{Joint Refinement of 3D Scenes and Poses} \label{method_BaRF}
% \subsubsection{3D Volume Rendering for 3D Radiance Field Reconstrution}
\noindent\textbf{Volume Rendering for Radiance Field Reconstruction.}
Based on the setting of neural volume rendering in NeRF, the radiance fields respective for geometry and appearance for a 3D scene are represented via two functions (implemented by MLPs): $F_\sigma: \mathbb{R}^3 \to \mathbb{R}^1$ and $F_c: \mathbb{R}^6 \to \mathbb{R}^3$, where $F_\sigma$ returns the volume density of an input 3D coordinate, while $F_c$ outputs the color at an input 3D coordinate given a 3D viewing direction. For rendering a pixel on 2D coordinate $u$ with its homogeneous form $\bar{u} = [u;1]^\top$, we first sample a sequence of $N$ 3D-coordinates $\{s_n\}_{n=1\cdots N}$ along the camera ray defined by the camera center $\vec{c} \in \mathbb{R}^3$ and the ray direction $\vec{d_u} = K^{-1}\bar{u}$, 
\begin{equation}
\begin{split}
    \medmath{ \left\{ s_n \right\}_{n=1\cdots N} =   s(\vec{c},\vec{d_u}) = \{ \vec{c} + t_n \cdot \vec{d_u} \}_{n=1\cdots N}},
\end{split}
\end{equation}
where $K$ is the intrinsic camera matrix and $\{t_n\}_{n=1\cdots N}$ are $N$ samples equidistantly distributed along the depth axis in between the near and far planes of the view frustum. The resultant color of the pixel is obtained by integrating through the density field $F_\sigma$ and color field $F_c$ using the volume rendering equation \cite{kajiya1984ray_volume_rendering,mildenhall2020nerf}, where we denote the \emph{discretized volume rendering intregral} by a function $\mathbf{V}$: 
\begin{equation}
\begin{split}
    \medmath{\mathbf{V}(F_\sigma, F_c, s(\vec{c},\vec{d_u)}) = \sum_{s_n \in s(\vec{c}, \vec{d})}^{} {T_n \cdot \alpha_n \cdot \mathbf{C}_n}},
\end{split}
\end{equation} 
where $T_n = exp(-\sum_{j =1}^{n} {\delta_j F_\sigma(s_j)})$ represents accumulated transmittance prior to $s_n$, $\alpha_n = 1-exp(-\delta_n F_\sigma(s_n))$ represents the opacity of sample $s_n$, and $\mathbf{C}_n =  F_c(s_n, \ \vec{d_u}\ )$ represents the color of sample $s_n$, and $\delta_j = \lVert s_j - s_{j-1} \rVert $ is the euclidean distance between two adjacent samples.
In the typical setting of NeRF, given a set of $L$ 2D-images $\mathbf{I} = \{I_1,\cdots, I_L\}$ with their corresponding camera poses $\mathbf{P} = \{P_1,\cdots, P_L\}$ $ \in \mathfrak{se}(3)$ Lie algebra (parametrizing rigid 3D transformation as $\mathfrak{se}(3)$ is a very common technique in robotics, here we follow the usage of \cite{lin2021barf}) as input, we aim to reconstruct the 3D scene represented by $F_\sigma^*$ and $F_c^*$, via minimizing the loss $\mathcal{L}_\text{rec}$ of 2D photometric reconstruction with the gradient-based optimization algorithm, in which 

% In the typical NeRF setting (with known camera pose) \cite{mildenhall2020nerf,chen2022tensorf}, we receive 2D images $\mathbf{I} = I_1, I_1, \cdots,  I_L$ along with the corresponding camera poses $\mathbf{P} = P_1, P_2, \cdots, P_L \in \mathfrak{se}(3)$ Lie algebra, where $L$ is the number of training views. We aim to reconstruct 3D scenes represented by $F_\sigma^*$ and $F_c^*$, by minimizing the loss of 2D photometric reconstruction with the gradient-based optimization algorithm. 
\begin{equation}
    \begin{split}
        \label{eq:recon3D}
        \medmath{
            \mathcal{L}_\text{rec}  (F_\sigma, F_c) = {\sum_{i=1}^L   \sum_{u\in \mathbf{U}} 
            {\lVert \mathbf{V}(F_\sigma, F_c, \mathcal{W}_\text{3d}(P_i, s(\vec{0}, \vec{d_u}))) - I_{iu}\rVert}}
        }, 
        %F_\sigma^*, & F_c^*  = \operatorname*{argmin}_{F_\sigma, F_c} L_{rec}(F_\sigma, F_c)
    \end{split}
\end{equation}
where $\mathbf{U}$ is the set of all possible 2D coordinates in the input images, $I_{iu} \in \mathbb{R}^3$ is the RGB color of pixel location $u$ on training image $I_i$, warping function $\mathcal{W}_\text{3d}(P, \_ ): \mathbb{R}^3 \to \mathbb{R}^3$ performs rigid 3D transformation parameterized by $P \in \mathfrak{se}(3)$ Lie algerbra, and $\mathcal{W}_\text{3d}(P, s(\vec{0}, \vec{d_u}))$ maps each sample 3D coordinate in canonical ray ($\vec{c} = \vec{0}$) into a 3D sample coordinate of camera ray with pose $P$. Note that this is an ill-posed reconstruction problem that suffers from shape-radiance ambiguity \cite{zhang2020nerf++}.

% \subsubsection{3D Joint Optimization.}
\noindent\textbf{3D Joint Optimization.}
When it comes to jointly estimating camera poses (where the camera poses $\mathbf{P}$ are also unknown) and learning scene representation \cite{lin2021barf,chng2022gaussian,chen2023local,heo2023robust}, the problem is even more ill-defined with the objective now being extended from Eq.~\ref{eq:recon3D} and defined as: 
\begin{equation}
    \label{eq:joint3D}
    \begin{split}
        \medmath{\mathcal{L}_\text{joint}  (F_\sigma, F_c, \mathbf{P}) = \sum_{i =1}^L  \sum_{u\in \mathbf{U}} 
        {\lVert \mathbf{V}(F_\sigma, F_c, \mathcal{W}_\text{3d}(P_i, s(\vec{0}, \vec{d_u}))) - I_{iu} \rVert}}.
        %F_\sigma^*, & F_c^*, \mathbf{P}^*  = \operatorname*{argmin}_{F_\sigma, F_c, \mathbf{P}}  L_{joint}(F_\sigma, F_c, \mathbf{P})
    \end{split}
\end{equation}
Such joint optimization is highly influenced by the structural bias of the underlying representation of $\{F_\sigma, F_c\}$, which we will conduct a pilot study with a simpler 1D case in Sec.~\ref{method_1D}.

\subsection{Gaussian Filter on 1D Signal Alignment} \label{method_1D}
Here we aim to analyze the effect of the signal spectrum (spectrum of $F_c$,  $F_\sigma$, and $\mathbf{I}$ in Eq.~\ref{eq:joint3D}) on the joint optimization process. We begin by reducing 3D joint optimization of camera pose and scene reconstruction into a simpler 1D counterpart of signal alignment. 

\noindent\textbf{1D Signal Alignment.}
Let us consider a target ground truth 1D signal $f_{GT}$ (assuming the signal to be continuous, bounded, and have finite support), which we aim to reconstruct and align with. We are given randomly translated versions $f_1$, $f_2$ of the ground truth signal $f_{GT}$, where $f_1 = \mathcal{W}_\text{1d}(f_{GT}, p_1), f_2 = \mathcal{W}_\text{1d}(f_{GT}, p_2)$ with having $\mathcal{W}_\text{1d}$ a signal translation operation defined as $\mathcal{W}_\text{1d}(f, p)(x) = f(x-p)$, and $p_1, p_2$ are the translation values.

Although the reconstruction is trivial in such a 1D setting, in order to mimic the case of 3D joint optimization, we attempt to estimate a signal $g$ as well as the  translation values $q_1$ and $q_2$ via adopting the iterative gradient-based optimization on the reconstruction loss.  
% Notice the similar structure of the formulation of $\mathcal{L}_\text{1d}$ Eq.~\ref{eq:1d_loss} and Eq.~\ref{eq:joint3D}.
\begin{equation}
    \label{eq:1d_loss}
    \begin{split}
        \medmath{\mathcal{L}_\text{1d}(g,q_1, q_2)}  & \medmath{ = \sum_{i\in [1,2]} \int \lVert \mathcal{W}_\text{1d}(g, q_i)(x) - f_i(x) \rVert^2 dx }\\
        &\medmath{=  \sum_{i\in [1,2]} \int \lVert g(x) - f_{GT}(x- p_i + q_i) \rVert^2 dx}.
        %g^*, q_1^*, q_2^* & = \operatorname*{argmin}_{g, q_1, q_2}  \mathcal{L}_\text{1d}(g, q_1, q_2)
    \end{split}
\end{equation}
Note that Eq.~\ref{eq:1d_loss} and Eq.~\ref{eq:joint3D} are analogous in terms of their structure/formulation, where the difference only lies in the dimensionality.
% Here the combination of $(q_1$, $q_2)$ is variable up to a translation; in fact,
And $\mathcal{L}_\text{1d}$ achieves the optimum whenever $q_1 - q_2 = p_1 - p_2$ and $g$ = $\mathcal{W}_\text{1d}(f_{GT}, p_1-q_1)$. Please check Figure~\ref{fig:1d_analysis}(a) for a simple visual representation of Equation~\ref{eq:1d_loss}, where $f_1$ and $f_2$ are connected to $g$ by the reconstruction loss $\mathcal{L}_\text{1d}$ (i.e blue arrows), whose gradients are used to update $g$ and the translation values $\{q_1, q_1\}$.

% \subsubsection{Correspondence between 1D Signal Alignment and 3D Joint Optimization.}
\noindent\textbf{Connection between 1D Signal Alignment and 3D Joint Optimization.}
The formulation of 1D signal alignment effectively simulates the ``local phenomenon'' of joint camera pose alignment and 3D scene reconstruction on a 2D cross-section:  
%(and also in the 2D image alignment case to be described in Section\ref{method_2D}). 
As shown in Figure~\ref{fig:1d_analysis}(c), where we consider two neighboring camera poses as well as a cross-section in the 3D space passing through both camera planes and intersecting with each camera plane on a projected straight line, the RGB color values on such projected lines correspond to the 1D shifted ground truth signals $f_1$, $f_2$ in Equation~\ref{eq:1d_loss}, and the value of the radiance field on the cross-section corresponds to reconstructed signal $g$ in Equation~\ref{eq:1d_loss}. 
Similar to the loss $\mathcal{L}_\text{1d}$ in Equation~\ref{eq:1d_loss}, the projected lines on the camera planes and the corresponding cross-section in the 3D radiance field are connected by the volume rendering function $V$ and reconstruction loss $\mathcal{L}_\text{joint}$ in Equation~\ref{eq:joint3D}. As a result, the complete 3D joint optimization can be intuitively viewed as simultaneously performing many 1D signal analyses on the superposition of all possible combinations of camera poses and cross-sections. 

% !!! ⚠️ ⚠️ ⚠️ THIS IMAGE NEED TO BE CLOSE TO THE MAIN TEXT for READABILITY  ⚠️ ⚠️ ⚠️ !!!!
% Combined Image : 1D Transfer Function and 1D-3D Correspondence  
\definecolor{ao(english)}{rgb}{0.0, 0.5, 0.0}

\begin{figure}[]
\centering
\includegraphics[width=\columnwidth]{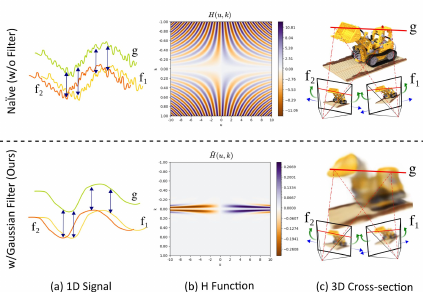} % Reduce the figure size so that it is slightly narrower than the column. Don't use precise values for figure width.This setup will avoid overfull boxes.
% \vskip -0.1in
\caption{
\textbf{Spectrum analysis and effect of Gaussian filtering on 1D signal alignment.}
% first row
% (a) Illustration 1D signal alignment with and without Gaussian Filtering. Notice the noisy signal on the left can easily get stuck in local optima. 
% (b) (\emph{Top}) Visualization of the spectral transfer function in Eq.~\ref{eq:1d_transfer}, notice the alternating sign in value of $H(u,k)$ when k leaves $0$, which causes the gradient-based optimization to move in the wrong direction if the spectrum of the signal contains too much energy in high-frequency domain. 
% (\emph{Bottom})The issue is mitigated by applying Gaussian filters on the signal, which effectively modulates the transfer function. 
% (c) The 1D alignment problem is similar to 3D joint optimization in Eq.~\ref{eq:joint3D}. The pose gradient signal that benefits pose refinement comes from the localized 1D alignment occurring in the cross-sections that are parallel to the camera error direction (the cross-section ({\color{red}{red lines in 3D scene}}) corresponds to the horizontal movement ({\color{blue}{blue arrows}}) and horizontal rotation ({\color{ao(english)}{green arrows}}).
(a) 1D signal alignment comparison: noisy signals can get trapped in local optima without Gaussian filtering. (b)(\emph{Top}) Visualization of $H(u, k)$ in Eq.~\ref{eq:1d_transfer}, which shows alternating signs as $k$ departs from $0$, causing misdirection in gradient-based optimization if there has too much high-frequency energy in the signal. (b)(\emph{Bottom}) Visualization of $\Tilde{H}(u, k)$ in Eq.~\ref{eq:1D_filter}, which is the modulated version of $H(u, k)$ with the help of Gaussian filter $\mathcal{N}$. (c) 1D alignment relates to 3D joint optimization in Eq.~\ref{eq:joint3D}, where effective pose refinement stems from the 1D alignment in specific cross-sections, with the {\color{red}{red lines in 3D scene}} correlating to horizontal shifts ({\color{blue}{blue arrows}}) and rotations ({\color{ao(english)}{green arrows}}).
}
% \vskip -0.1in
\label{fig:1d_analysis}
\end{figure}

% \subsubsection{Spectrum Analysis and Effect of Gaussian Filtering on 1D Signal Alignment.}
\noindent\textbf{Spectrum Analysis and Effect of Gaussian Filtering on 1D Signal Alignment.}
% Due to paragraph space limitation, we move detailed formulations and analysis to the supplementary material and only outline a sketch below: 
% \begin{itemize}
    % \item 
First we transform the problem into a simpler form with a assumption that is reflected by the fast convergent property of voxel grids (cf. our supplement for detailed derivation of the theorem): 
\begin{theorem}
If we assume rapid convergence of signal $g$ (which means $g$ achieves local  optima $g^*$ w.r.t current $q_1, q_2$ whenever we update $q_1, q_2$.), we find that the problem in Eq.\ref{eq:1d_loss} is equivalent to pure alignment between two ground-truth signals, that is
 \begin{equation}
    \label{eq:align_1D}
    \begin{aligned}
        \medmath{\mathcal{L}_\text{1d}(g, q_1, q_2) }&= \medmath{\mathcal{L}_\text{1d}(g^*, q_1, q_2) }\\ &= \medmath{ \mathcal{L}_\text{1d}(u)  =  \int \lVert f_{GT}(x) - f_{GT}(x+ u) \rVert^2 dx},
    \end{aligned}
\end{equation}
where $u = (p_1-p_2) - (q_1-q_2)$ is the shift between two ground truth signals, which has an initial value of $p_1 - p_2$
\end{theorem}
We aim for $u$ to reach $0$ with gradient-based optimization.

    % \item 
    Next, by analyzing the relationship between $f_{GT}$ and the optimization gradient ${d \over d u}\mathcal{L}_\text{1d}$ in terms of their spectral properties, we get the following result (cf. our supplement for detailed derivation of the theorem):
    \begin{theorem}
        \begin{equation}
        \label{eq:1d_transfer}
        \begin{aligned}
            \medmath{{d \over d u}\mathcal{L}_\text{1d}} & \medmath{=   \int \lVert \ \mathfrak{F}[{f_{GT}]} \ \rVert^2  \cdot H(u,k) \ dk},
              % \text{where } & H(u,k)  = 4 \pi k \  sin( 2 \pi k u)
        \end{aligned}
        \end{equation}
    where $H(u,k)  = 4 \pi k \  sin( 2 \pi k u)$, $\mathfrak{F}[f_{GT}]$ is Fourier transform 
    % \cite{bracewell1986fourier} 
    of $f_{GT}$, and $k$ is the wavenumber in frequency domain. 
    \end{theorem}
    Particularly, we are interested in the sign of ${d \over d u}\mathcal{L}_\text{1d}$ which determines the direction of our iterative optimization. We plot the value of $H(u,k)$ in Fig.~\ref{fig:1d_analysis}(b)(\emph{Top}), where we can observe that the sign of $H$ is well-behaved when the magnitude of $k$ is small (here well-behaving means the direction of the gradient is able to let $u$ descend to $0$, i.e., being positive when $u>0$ and negative when $u<0$). However, when $k$ increases, the sign of $H$ quickly begins to alternate, and the magnitude increases, which causes the gradient to be large and noisy. Hence high-frequency signals with a spreading spectrum can easily lead the optimization process to get stuck in the local optima.
    
    % \item 
    To this end, we demonstrate that applying a Gaussian filter on the signal $f_{GT}$ effectively mitigates the sign-alternating issue of the original $H$ function. Specifically, we show that filtering the input signal is equivalent to modulating $H$ by a Gaussian window (cf. our supplement for derivation):
    \begin{theorem}
        Let  $\Tilde{\mathcal{L}}_\text{1d}$ denotes $\mathcal{L}_\text{1d}$ calculated with Gaussian convoluted signal $\mathcal{N} \ast f_{GT}$, and $\mathfrak{F}[\mathcal{N}]$ denotes the Fourier transform of the Gaussian kernel $\mathcal{N}$, then we have 
        \begin{equation}
        \label{eq:1D_filter}
        \begin{aligned}
            \medmath{{d \over d u}  \Tilde{\mathcal{L}}_\text{1d} }& \medmath{=   \int \lVert \ \mathfrak{F}[f_{GT}] \ \rVert^2 \cdot \Tilde{H}(u,k) \ dk},
            % \text{where } &  H'(u,k) = \rVert \ F[\ N \ ] \lVert ^2 \cdot H(u,k) 
        \end{aligned}
        \end{equation}
        where $\Tilde{H}(u,k) = \rVert \ \mathfrak{F}[\mathcal{N}] \ \lVert ^2 \cdot H(u,k).$
    \end{theorem}
    In Fig.~\ref{fig:1d_analysis}(b)(\emph{Bottom}), we plot the modulated $\Tilde{H}(u,k)$, with observing that the misbehave region is suppressed (note that we set the variance of $\mathcal{N}$ to $4$ here). The gradient descent will likely converge to $u=0$ once the initial magnitude of $u$ is less than $6.0$. The region where ${d \over du} \Tilde{\mathcal{L}}_\text{1d}$ does well-behave is \emph{quasi-convex} and is guaranteed to converge to global optima given suitable learning rate that prevents us from getting stuck on saddle points.
    % \item 
    Our analysis agrees with the motivation behind the coarse-to-fine training schedule of \cite{lin2021barf} and \cite{heo2023robust}. Specifically, observing that the well-behaved region in $H(u,k)$ grows wider as $u$ approaches $0$ (cf. Fig.~\ref{fig:1d_analysis}(b)(\emph{Top})), which means that we can loosen the filtering strength of Gaussian kernel as $u$ approaches $0$, leading to larger and more accurate gradient. 
% \end{itemize}

\subsection{2D Planar Image Alignment} \label{method_2D}

In addition to the 3D joint optimization problem, previous works \cite{lin2021barf,chng2022gaussian} also consider a 2D image patches alignment task as a simpler example of joint optimization, in which there are $L_\text{2d}$ overlapping image patches $ \mathbf{I_\text{2d}} =  \{I_1, \cdots, I_{L_\text{2d}} \}$ cropped from a single ground truth image $I_{gt}$ before being transformed by 2D homography. The homography transforms are parameterized by $\mathbf{P_\text{2d}} = \{P_1, \cdots, P_{L_\text{2d}} \} \in \mathfrak{sl}(3)$ and initialized as $\vec{0}$ (here we also follow from \cite{lin2021barf} the usage Lie algrebra to parameterize 2D homography transform). Analogously to Equation \ref{eq:joint3D}, our objective is to jointly optimize the 2D image content $F_\text{2d} : \mathbb{R}^2 \to \mathbb{R}^2$ and per-patch homography warps $\mathbf{P_\text{2d}}$ by the reconstruction loss. Joint optimization can be formulated as:

\begin{equation}
 \label{eq:joint2D}    
    \begin{split}
        \medmath{\mathcal{L}_\text{2d}  (F_\text{2d}, \mathbf{P_\text{2d}}) =  \sum_{i = 1}^{L}
        { \sum_{ u \in  \mathbf{U_\text{2d}}}  {}}
        {\lVert F_\text{2d}(\mathcal{W}_\text{2d}(P_i, u)) - I_{iu}\rVert}^2},
        %{F_\text{2d}}^* &, \  \mathbf{P_\text{2d}}^* = \operatorname*{argmin}_{F,\mathbf{P}} {L_\text{2d}(F,\mathbf{P})}
    \end{split}
\end{equation}
where $\mathbf{U_\text{2d}}$ is the set of all possible 2D coordinates in the image patches, $I_{iu}$ is the color of pixel at location $u$ on input image patch $I_i$, warp function $\mathcal{W}_\text{2d}(P_i, \_ ): \mathbb{R}^2 \to \mathbb{R}^2$ performs 2D homography transformation parameterized by $P_i \in \mathfrak{sl}(3)$ Lie algebra, and $\mathcal{W}_\text{2d}(P_i, u)$ maps 2D coordinate $u$ on $I_{gt}$ into a transformed 2D coordinate on patch $I_i$. Notice the strong structural correspondence among Eq.~\ref{eq:1d_loss} (1D alignment), Eq.~\ref{eq:joint2D} (2D alignment), and Eq.~\ref{eq:joint3D} (3D alignment), the three problems share similar computational property.

We parameterize $F_\text{2d}$ by a 2D decomposed low-rank tensor $\Tau_\text{2d} \in \mathbb{R}^{h \times w}$, where $w,h$ are the dimensions of the image. Motivated by our analysis in Section \ref{method_1D}, we filter $\Tau_\text{2d}$ with 2D Gaussian kernel to avoid overfitting. 
% \begin{equation}
%     \label{2d:tensor}
%     \begin{aligned}
%         F_\text{2d}(x,y) & = (N_\text{2d} \ast \Tau_\text{2d})(x,y) = (N_\text{2d} \ast (\sum^{R}_{r=1} \mathbf{v}^X_r \otimes \mathbf{v}^Y_r))(x,y)
%     \end{aligned}
% \end{equation}
\begin{equation}
    \label{2d:tensor}
    \begin{split}
        % \medmath{F_\text{2d}(x,y) = (N_\text{2d} \ast \Tau_\text{2d})(x,y) = (N_\text{2d} \ast (\sum^{R}_{r=1} \mathbf{v}^X_r \otimes \mathbf{v}^Y_r))(x,y)},
        \medmath{F_\text{2d}(\textbf{x}) = (\mathcal{N}_\text{2d} \ast_\text{2d} \Tau_\text{2d})(\textbf{x}) = (\mathcal{N}_\text{2d} \ast_\text{2d} (\sum^{R}_{r=1} \mathbf{v}^X_r \otimes \mathbf{v}^Y_r))(\textbf{x})},
    \end{split}
\end{equation}
where $\textbf{x} \in \mathbb{R}^2$ is 2D pixel coordinates, $\mathcal{N}_\text{2d}$ is 2D gaussian kernel, $\ast_\text{2d}$ is the convolution operator, and $\otimes$ denotes outer product between the 1D vector components $\mathbf{v}^X_r \in \mathbb{R}^w, \mathbf{v}^Y_r \in \mathbb{R}^h$. ``$(\textbf{x})$'' at the end of the expressions means bilinearly interpolating the preceding discrete 2D volume with continuous coordinate \textbf{x}. Our method outperforms the na\"ive tensor method and previous methods \cite{lin2021barf,chng2022gaussian},  experiment results are shown at Sec.~\ref{exp:2d_result}.

The width of Gaussian kernel $\mathcal{N}_\text{2d}$ is controlled by an exponential coarse-to-fine training schedule that changes continuously (cf. our supplement for details of such kernel schedule). In order to support continuous changing width on a discrete Gaussian kernel, the kernel is generated by the following rule: 
    \begin{equation}
        \label{eq:kernel}
        \begin{aligned}
            \medmath{\mathcal{N}_\text{1d}(\sigma) }& \medmath{=} 
                \begin{cases}
                    \medmath{\bigoplus^{L_\mathcal{N} / 2}_{x=-{L_{\mathcal N} /2}}  min(1,{1 \over \sqrt{2 \pi}\sigma} e^{- {x^2 \over 2 \sigma^2}}) }&\medmath{\text{if }\sigma > 0.0001}\\
                    \medmath{\bigoplus^{L_\mathcal{N} / 2}_{x=-{L_{\mathcal N}/2}} \delta[x]} &\medmath{\text{otherwise}},
                \end{cases}\\
            \medmath{\mathcal{N}_\text{2d}(\sigma)}& \medmath{= \mathcal{N}_\text{1d}(\sigma) \otimes \mathcal{N}_\text{1d}(\sigma)},
        \end{aligned}
    \end{equation}
    % \begin{equation}
    %     \label{eq:kernel}
    %     \begin{aligned}
    %         N_\text{1d}(\sigma) & = 
    %             \begin{cases}
    %                 \bigoplus^{\mathbf{N}}_{x=1}  min(1,{1 \over \sqrt{2 \pi}\sigma} e^{- {x^2 \over 2 \sigma^2}}) &\text{if }\sigma > 0.0001\\
    %                 \delta[x] &\text{otherwise},
    %             \end{cases}\\
    %         N_\text{2d}(\sigma) & = N_\text{1d}(\sigma)^{\top} \otimes N_\text{1d}(\sigma),
    %     \end{aligned}
    % \end{equation}
where $L_\mathcal{N}$ is the size of the discrete kernel, 1D kernel $\mathcal{N}_\text{1d}(\sigma) \in \mathbb{R}^{L_\mathcal{N}}$ is discretely sampled from continuous Gaussian distribution and clamped to a max value of $1.0$ before being concatenated into a vector by $\oplus$ operator. To avoid numerical instability, when $\sigma < 0.001$,  we assign $\mathcal{N}_\text{1d}(\sigma)$ to be discrete impluse function $\delta$. 2D kernel $\mathcal{N}_\text{2d}(\sigma) \in \mathbb{R}^{L_\mathcal{N}\times L_\mathcal{N}}$ is generated by outer product of two 1D kernels.

\subsection{Decomposed Low-Rank Tensor}  \label{method_TensoRF} 
This section describes the decomposed low-rank tensor proposed by TensoRF~\cite{chen2022tensorf} which is the scene representation that our proposed method is built upon. While there are two different types of tensor decomposition considered in~\cite{chen2022tensorf}: CP-decomposition and VM-decomposition, in our discussion we mainly focus on \emph{VM-decomposition}, although our method is also naturally applicable to CP-decomposition.

To represent the 3D density field $F_\sigma$, we store the information in a 3D tensor $\Tau_\sigma \in \mathbb{R}^{I\times J \times K}$, in which now $F_\sigma$ is defined simply as component-wise interpolation of $\Tau_\sigma$.
\begin{equation}
\label{eq:sigma_tensor}
     \medmath{\Tau_\sigma = \sum_{r=1}^{\mathbf{R}} {\mathbf{v}^X_{\sigma, r}} \otimes \mathbf{M}^{Y,Z}_{\sigma,r} + 
                                       {\mathbf{v}^Y_{\sigma, r}} \otimes \mathbf{M}^{X,Z}_{\sigma,r} + 
                                       {\mathbf{v}^Z_{\sigma, r}} \otimes \mathbf{M}^{X,Y}_{\sigma, r}},
\end{equation}
where $\mathbf{R}$ is the number of components in the decomposition, $(\mathbf{v}^X_r , \mathbf{v}^Y_r, \mathbf{v}^Z_r) \in (\mathbb{R}^I, \mathbb{R}^J, \mathbb{R}^K)  $ are 1D vector-components for axes $(X, Y, Z)$ repectively, $(\mathbf{M}^{Y,Z}_r , \mathbf{M}^{X,Z}_r, \mathbf{M}^{X,Y}_r) \in  (\mathbb{R}^{J \times K},  \mathbb{R}^{I \times K}, \mathbb{R}^{I \times J})$ are 2D matrix-components for axes $(Y\textnormal{-}X, X\textnormal{-}Z, X\textnormal{-}Y)$ repectively, operator $\otimes$ denotes the outer product between vector and matrix. 

To represent the 3D color field $F_c$, the information $\Tau_c(\textbf{x}) \in \mathbb{R}^G$ queried from 3D feature tensor $\Tau_c \in \mathbb{R}^{I\times J \times K \times G}$ is decoded by a small MLP $S$ into RGB color value ($G$ is the input feature dimension of $S$). The implementation can be formulated as
\begin{equation}
\label{eq:color_tensor}
\begin{aligned}
    & \medmath{F_c(\textbf{x},\vec{d})  = \mathbf{S}( \Tau_c(\textbf{x}), \vec{d})}\\
        &\medmath{\Tau_c = \medmath{\sum_{r=1}^{\mathbf{R}} {\mathbf{v}^X_{c,r}} \otimes \mathbf{M}^{Y,Z}_{c,r} \otimes \mathbf{b}^X_r + }
        }\\ 
        & \medmath{\quad \quad \quad \medmath{{\mathbf{v}^Y_{c,r}} \otimes \mathbf{M}^{X,Z}_{c,r} \otimes \mathbf{b}^Y_r + 
                {\mathbf{v}^Z_{c,r}} \otimes \mathbf{M}^{X,Y}_{c,r} \otimes \mathbf{b}^Z_r }
                }.
\end{aligned}
\end{equation}
$\Tau_c(\textbf{x})$ denotes the component-wise linear-interpolation of tensor volume $\Tau_c$ on 3D coordinate \textbf{x}. $\vec{d}$ is the viewing direction of the current ray. $\mathbf{v}_{c,r}$ and $\mathbf{M}_{c,r}$ have the same shape as their $\mathbf{v}_{\sigma,r}$ and $\mathbf{M}_{\sigma,r}$ counterparts, $\mathbf{b}^X_r, \mathbf{b}^Y_r, \mathbf{b}^X_r \in \mathbb{R}^G$ are feature components to expand the feature axis of $\Tau_c$.

% \subsection{Separable Component-Wise Convolution for Efficient Spectrum Control on 3D Decomposed Low-Rank Tensor} \label{method_3D}
\subsection{Separable Component-Wise Convolution} \label{method_3D}
As theoretically analyzed in Sec.~\ref{method_1D} and empirically shown in Fig.~\ref{fig:naive}(a), na\"ively applying low-rank decomposed tensor (which lacks internal bias that limits the spectrum of learned signal, hence corresponds to the top raw of Fig. \ref{fig:1d_analysis}) to joint camera pose optimization results in suboptimal reconstruction quality and inaccurate poses. Therefore, we propose to limit the spectrum of the radiance field $F_\sigma$ and $F_c$ with a coarse-to-fine training schedule. 

If we na\"ively convolve the 3D Gaussian kernel with our 3D volume $\Tau_\sigma$, (as in the 2D planar case of Eq.~\ref{2d:tensor}), we would have to reconstruct the whole 3D tensor before applying convolution, destroying the space compression advantage of decomposed low-rank tensor, see Eq.~\ref{eq:brute-force}.

\begin{equation}
    \label{eq:brute-force}
    \begin{aligned}
        \medmath{F_\sigma(x,y,z) = (\mathcal{N}_{\text{3d}} \ast_\text{3d} \Tau_\sigma)(x,y,z)},
    \end{aligned}
\end{equation}
where $\ast_{\text{3d}}$ denotes 3D convolution, $\mathcal{N}_{\text{3d}}$ is the 3D Gaussian filter defined by $\mathcal{N}_{1d} \otimes \mathcal{N}_{\text{2d}}$. Under this setting, the time complexity and the space complexity are $O(I\cdot J\cdot K\cdot L_\mathcal{N}^3)$ and $O(I\cdot J \cdot K)$ respectively, where $L_\mathcal{N}$ is the size of 3D Gaussian kernel in each dimension.

%% This part is optional
\begin{comment}
In order to preserve the space compression benefit offered by decomposed low-rank tensor, we can perform convolution locally on the queried 3D coordinate (i.e., we densely query $N^3$ neighboring values out of the original tensor, then calculate the sum of the product with the 3D Gaussian kernel). The time complexity and space complexity will be $O(N^3 \times S))$, where $S$ is the number of samples queried. See Equation \ref{eq:naive_3d_conv} below for more details.

\begin{equation}
\label{eq:naive_3d_conv}
\begin{aligned}
    \medmath{F_\sigma(x,y,z) = \sum_{i=-\frac{N}{2}}^{\frac{N}{2}} \sum_{j=-\frac{N}{2}}^{\frac{N}{2}} \sum_{k=-\frac{N}{2}}^{\frac{N}{2}} \Tau_\sigma(x-i,y-j,z-k) \cdot N_{\text{3d}}(i,j,k) }
\end{aligned}
\end{equation}

However, Equation \ref{eq:naive_3d_conv} is still too slow for practical use.
\end{comment}
 
To achieve computationally efficient convolution on the 3D decomposed low-rank tensor volume, we perform our proposed \emph{separable component-wise convolution}, by taking advantage of the following identity (whose correctness will be proven in the supplementary material). 
\begin{theorem}
\begin{equation}
    \label{eq:distribute_sigma}
    \begin{aligned}
         \Tilde \Tau_\sigma  = \medmath{\sum_{r=1}^{\mathbf{R}} 
                                       {\mathbf{\Tilde v}^X_{\sigma, r}} \otimes {\mathbf{\Tilde M}^{Y,Z}_{\sigma,r}} + 
                                       {\mathbf{\Tilde v}^Y_{\sigma, r}} \otimes {\mathbf{\Tilde M}^{X,Z}_{\sigma,r}} + 
                                       {\mathbf{\Tilde v}^Z_{\sigma, r}} \otimes {\mathbf{\Tilde M}^{X,Y}_{\sigma,r}}},
    \end{aligned}
\end{equation}
\end{theorem}
where $\Tilde \Tau_\sigma = (\mathcal{N}_{\text{3d}} \ast_{\text{3d}} \Tau_\sigma)$ denotes the 3D Gaussian convoluted tensor volume, ${\mathbf{\Tilde v}_{\sigma, r}} = (\mathcal{N}_{\text{1d}} \ast_{\text{1d}} \mathbf{\Tilde v}_{\sigma,r})$ denotes the 1D Gaussian convoluted vector component, and ${\mathbf{\Tilde M}_{\sigma,r}} = (\mathcal{N}_{\text{2d}} \ast_{\text{2d}} \mathbf{\Tilde M}_{\sigma, r})$ denotes the 2D Gaussian convoluted matrix component. 
In other words, \textbf{the 3D convoluted tensor can be expressed as the composition of individually convoluted components}, which allows us to distribute the 3D Gaussian convolution across the individual components of the decomposed low-rank tensor. Similar to Sec.~\ref{method_TensoRF}, the value of the density field is component-wised linearly sampled from the Gaussian convoluted components, i.e., $\Tilde{F_\sigma}(\textbf{x}) = \Tilde{\Tau_\sigma}(\textbf{x})$. 
Similarly, the spectral restricted version of the color field $F_c$ can be obtained as 
\begin{equation}
    \label{eq:distribute_color}
    \begin{aligned}
        & \medmath{\Tilde{F_c}(\textbf{x},\vec{d})  = \mathbf{S}( \Tilde \Tau_c(\textbf{x}), \vec{d})}\\
        & \medmath{\Tilde \Tau_c = \sum_{r=1}^{\mathbf{R}} { \mathbf{\Tilde v}^X_{c,r}} \otimes \mathbf{\Tilde M}^{Y,Z}_{c,r} \otimes \mathbf{b}^X_r + }\\ 
        & \quad \quad \quad 
        \medmath{
            {\mathbf{\Tilde v}^Y_{c,r}} \otimes \mathbf{\Tilde M}^{X,Z}_{c,r} \otimes \mathbf{b}^Y_r + 
            {\mathbf{\Tilde v}^Z_{c,r}} \otimes \mathbf{\Tilde M}^{X,Y}_{c,r} \otimes \mathbf{b}^Z_r 
        }.
    \end{aligned}
\end{equation}
With \emph{separable component-wise convolution}, the time complexity required is $O(I\cdot J \cdot L_\mathcal{N} + J \cdot K \cdot L_\mathcal{N} + K \cdot I \cdot L_\mathcal{N})$ for computing convoluted components (assuming that we separate 2D Gaussian convolution on matrix components into 1D Gaussian convolutions), and $O(\mathbf{R})$ for each query sample (same as the original decomposed tensor in \cite{chen2022tensorf}), drastically reducing the computation required for filtering 3D radiance fields $F_\sigma$ and $F_c$.

We stress here that our proposed component-wise convolution is \textbf{different} from traditional technique of separated kernel convolution in signal processing literature, in the sense that the common separated kernel technique only separates the 3D kernel without utilizing the separability of the input signal itself, and hence requires sequentially performing three 1D convolution operation on 3D volume, the time complexity of traditional technique would be $O(I \cdot J \cdot K \cdot L_\mathcal{N})$, and also requires a 3-dimensional memory with space complexity of $(I \cdot J \cdot K)$ to store convolution result.

\subsection{Techniques for Increasing Pose Robustness} \label{method_Robust}
Here we summarize our improvements on na\"ive decomposed low-rank tensors that improve joint camera pose optimization and radiance field reconstruction. 
\subsubsection{Coarse-to-Fine 3D schedule.}
Using efficient 3D convolution algorithm in Sec. \ref{method_3D}. During training, we apply a coarse-to-fine schedule on the 3D radiance field $\Tilde F_\sigma, \Tilde F_c$ by controlling the kernel parameter ($\sigma$ of Eq.~\ref{eq:kernel}) of the Gaussian kernel, which is exponentially reduced to 0 at 10k iterations and remains 0 afterward (for detailed settings of $\sigma$, please refer to the supplement).

%%However, a single fixed kernel schedule is not robust enough to handle various input scenes. Therefore, we further proposed other techniques to further increase the robustness of joint optimization on decomposed low-rank tensors.

\subsubsection{Smoothed 2D Supervision.}
    Inspired by the analysis in Sec.~\ref{method_1D}, we discovered that blurring the 2D training image with a parallel set of scheduled 2D Gaussian kernels also helps the joint optimization.  On the one hand, smoothed supervision images produce smoothed image gradients and stabilize the camera alignment. On the other hand, smoothed training image also helps to restrict the spectrum of the learned 3D scene. The Gaussian schedule for smoothing 2D training images is similar to that of the 3D radiance fields .
\subsubsection{Randomly Scaled Kernel Parameter and Edge Guided Loss.}
\begin{figure*}[!ht]
    \centering
    \includegraphics[width=1.0 \textwidth]{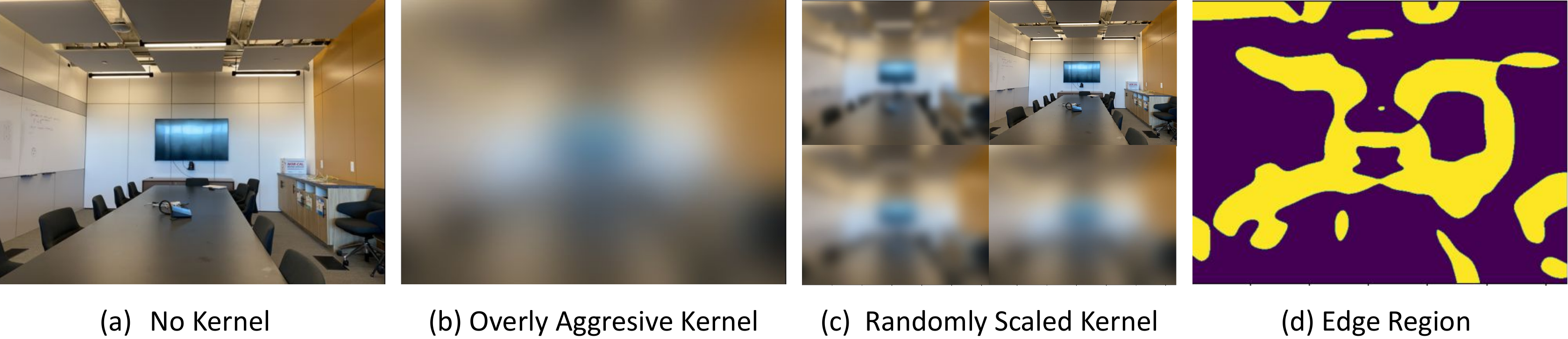}
    \caption{\textbf{Visualization of 2D Randomly Sampled Kernel and Edge Guided Loss}. (a) Input supervision without kernel. Joint optimization using unblurred images easily overfit to high-frequency noises (b) Input supervision blurred by an overly aggressive kernel. Notice that the edge information is largely destroyed by the blurring process, resulting in weak and noisy gradients, causing the poses to drift around easily. (c) Same input supervision blurred by four randomly scaled kernels. We empirically found that mixing different filtering strengths results in a more robust joint optimization. (d) We select edge area of a blurred image by Sobel filter with a threshold set to 1.25x of the average value of the filtered edge-strength map.}
    \label{fig:randomscale_edge_vis}
\end{figure*}

    From the previous spectral analysis in Sec. \ref{method_1D}, one may have the impression that a larger kernel leads to stronger modulation, and hence always results in more robust pose registration. However, this is not always true, because the magnitude of $H(u,k)$ decreases linearly as $k$ approaches $0$. Notice that in Fig. \ref{fig:1d_analysis}(b) the magnitude of modulated $\Tilde{H}$ is weaker than that of $H$, which means that $\frac{d}{du}\Tilde{\mathcal{L}}_\text{1d}$ is weaker than $\frac{d}{du}\mathcal{L}_\text{1d}$ and therefore is more easily influenced by noise. In the 3D case, this \emph{\textbf{weak and noisy gradient problem}} caused by overly aggressive filtering corresponds to the excessive blur effect that destroys important edge signals in the training images, causing pose alignment to fail. See Fig. \ref{fig:randomscale_edge_vis}(b) for a visualization of the image blurred by an over-strength kernel, in which the thin edge information is eliminated, causing the camera pose to randomly drift.

Based on the effect of \emph{weak and noisy gradient problem}, when applying only \emph{coarse-to-fine 3D schedule} and \emph{smoothed 2D supervision},  we found that it is insufficient to use a single-size kernel on different real-world scene structures (in which the same kernel may be overly aggressive in one scene, but overly gentle in another scene). Therefore, we introduce \textbf{\emph{randomly scaled kernel}}, which randomly scales the kernel by a factor uniformly sampled from $[0,1]$. Random scales are sampled independently among 3D Gaussian kernels (for the radiance field) and 2D Gaussian kernels (for training images), allowing combinations of different-sized kernels to guide the joint optimization. See Fig. \ref{fig:randomscale_edge_vis}(c) for a visualization of the same input image filtered by a range of randomly sampled kernels. We observe that the training schedule becomes more robust when we alternate between these randomly sampled kernel scales.

Another way to mitigate the weak and noisy gradient problem is the \textbf{\emph{edge guided loss}} , in which we increase the learning rate by 1.5x (and hence amplify the gradient signal) on the pixels in the edge region, from which the learning signal for pose alignment mainly comes. See visualization in Fig. \ref{fig:randomscale_edge_vis} (d), where we color the edge area that is detected using the Sobel filter \cite{kanopoulos1988design_sobel_filter} on the filtered 2D images in yellow. Edge-guided rendering loss helps the joint optimization focuses more on the edge area of the training images, resulting in more robust pose optimization. Empirically we apply this edge-guided scale alternately on every other training iteration.

%% file: CameraReady/LaTeX/exp.tex
\section{Experiments}
Although our method is applicable to various decomposed low-rank tensor implementations, in this section, we validate our proposed method using TensoRF~\cite{chen2022tensorf} with inaccurate or unknown camera poses. 

We evaluate our proposed method against three previous works \textbf{BARF} \cite{lin2021barf}, \textbf{GARF} \cite{chng2022gaussian}, and \textbf{HASH} \cite{heo2023robust}. Since the implementation of GARF and HASH are unavailable, we directly use the results reported in their paper for comparison.
We compare these methods on the \textbf{planar image alignment} task and novel view synthesis task on \textbf{NeRF-Synthetic} and \textbf{LLFF} dataset. 
% Due to the space limit, w
We provide detailed implementation details and experimental setup in the supplementary material.

\subsection{Results}
\subsubsection{Planar Image Alignment (2D).}\label{exp:2d_result}
\begin{figure}[t]
    % \vskip 0.2in
    \begin{center}
    \centerline{\includegraphics[width=1.0\columnwidth]{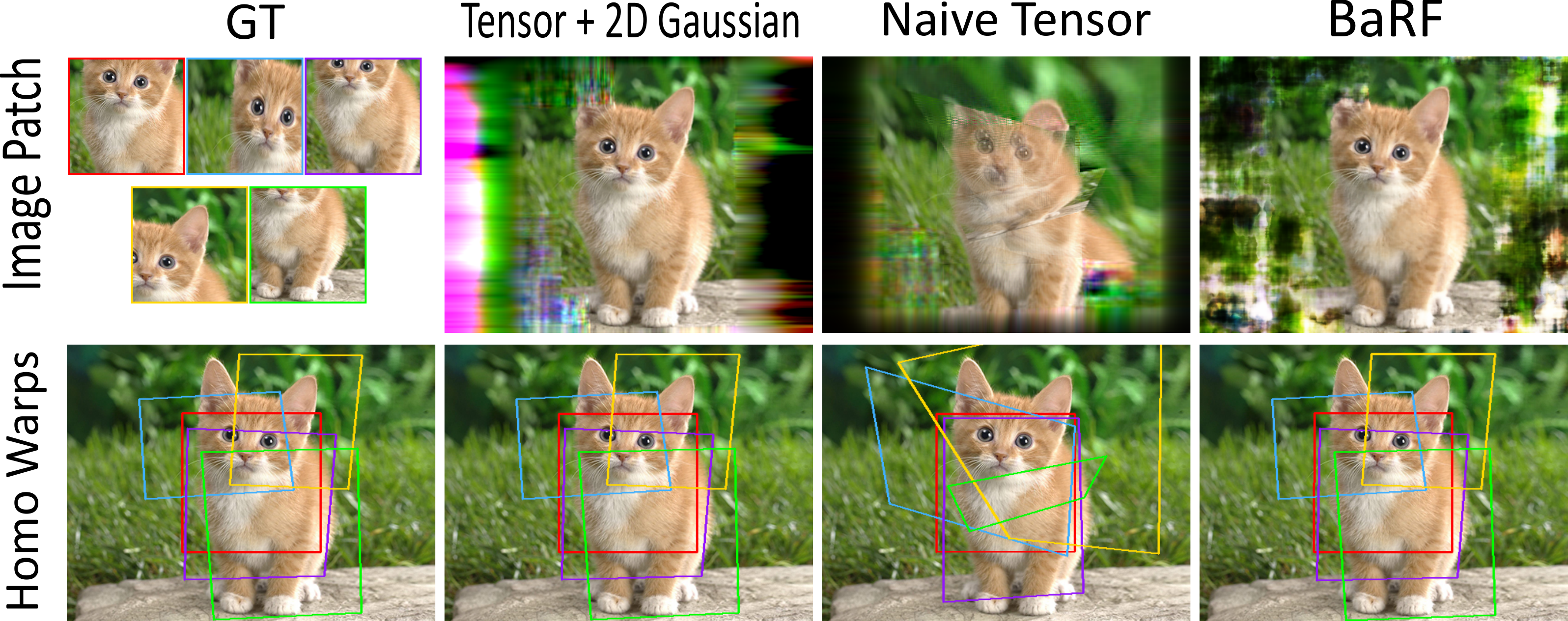}}
    % \vskip -0.05in
    % \vskip -0.15em
    \caption{\textbf{Qualitative comparisons of the 2D image patch alignment.} \emph{2D TensoRF + 2D Gaussian} successfully registers accurate warping parameters, verifying the analysis of Gaussian filtering on joint optimization.
    % in Section \ref{method_1D}. See \ref{exp:2d_result} for quantitative results.
    % ~\yulunliu{2D TensoRF instead of Tensor?}~\boyu{fixed} 
    }
    \label{fig:2d_qualitative}
    \end{center}
    % \vskip -0.2in
    % \vskip -0.2in
\end{figure}

In Fig.~\ref{fig:2d_qualitative} we compare our method (i.e., 2D TensoRF + 2D Gaussian) with na\"ive 2D TensoRF implementation~\cite{chen2022tensorf} and BARF \cite{lin2021barf}. Quantitative results are reported in Tab.~\ref{tab:results_2d}, including $\mathfrak{sl}(3)$ warp error and patch PSNR. These results demonstrate the effectiveness of Gaussian filtering in joint optimization, verifying the analysis in Sec.~\ref{method_1D}.
\begin{table}[t]
\centering
\footnotesize
% \resizebox{1.0\columnwidth}{!}{%
 \begin{tabular}{l|c|c}
                \toprule
                Methods & $\mathfrak{sl}(3)$ error $\downarrow$ & patch PSNR $\uparrow$ \\
                \midrule
            BARF & 0.0105 & 35.19 \\
            Na\"ive 2D TensoRF   & 0.5912 & 20.80 \\
            2D TensoRF + 2D Gaussian & \textbf{0.0023} &  \textbf{40.70} \\   
    \bottomrule 
\end{tabular}
% }
% \vskip -0.1in
\caption{\textbf{Quantitative results of planar image alignment.}}
\label{tab:results_2d}
\end{table}
%  \begin{tabular}{l|c|c}
%                 \toprule
%                 Methods & $\mathfrak{sl}(3)$ error $\downarrow$ & patch PSNR $\uparrow$ \\
%                 \midrule
%             BARF & 0.0105 & 35.19 \\
%             Tensor Na\"ive  & 0.5912 & 20.80 \\
%             Tensor + 2D Gaussian & \textbf{0.0023} &  \textbf{40.70} \\   
%     \bottomrule 
% \end{tabular}
% Figure: 2D Planar Image Alignement Results

\begin{figure}[t]
\centering
\resizebox{\columnwidth}{!}{%
\includegraphics[width=0.95\columnwidth]{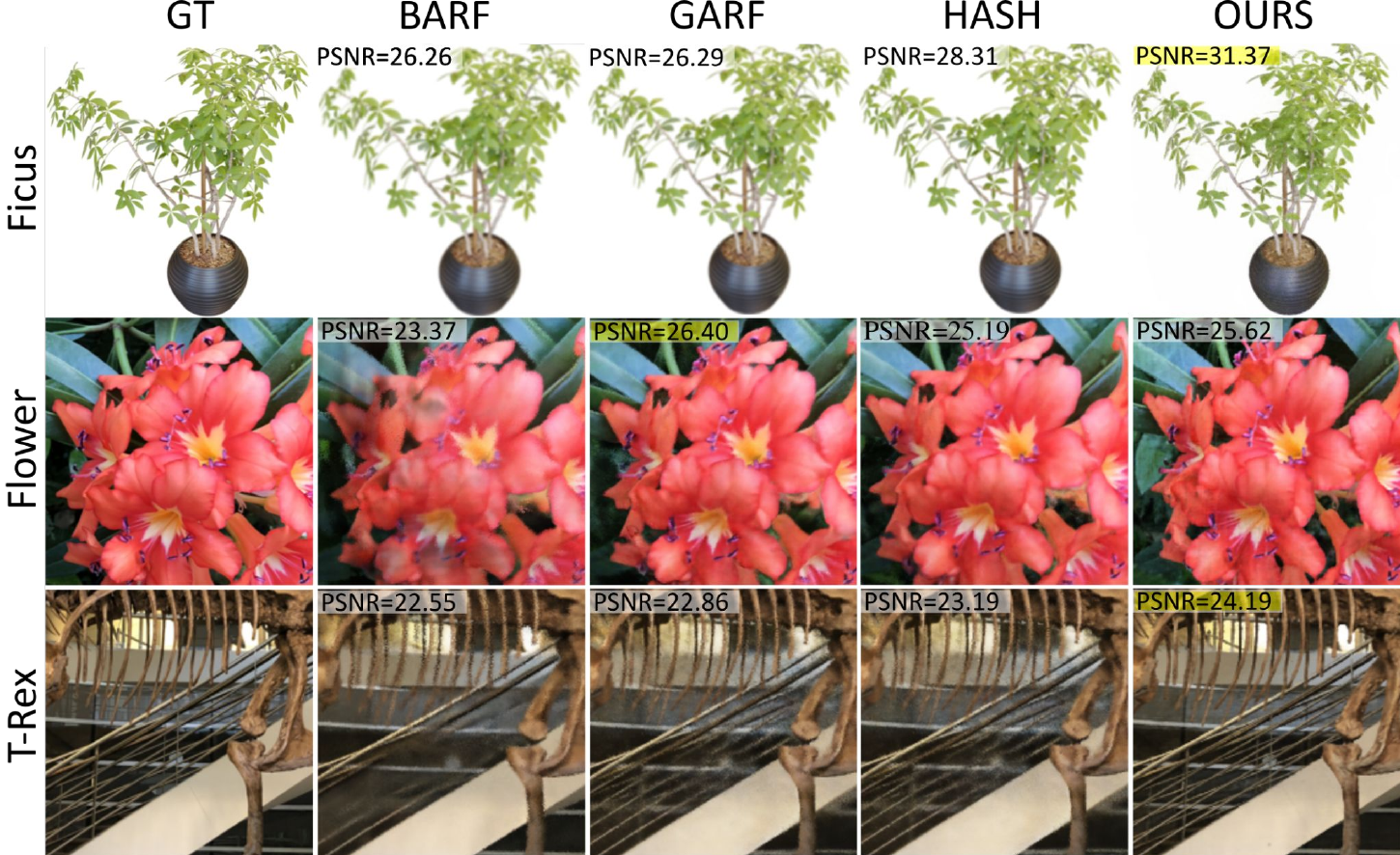}
}
% \vskip -0.05in
\caption{\textbf{Visual comparisons of novel view synthesis.}}
% \vskip -0.1in
\label{fig:model_comparison}
\end{figure}

\input{CameraReady/LaTeX/Tables/synthetic}

\input{CameraReady/LaTeX/Tables/llff}

\subsubsection{NeRF (3D): Synthetic Object \& Real World Objects.}\label{exp:blender} \label{exp:llff}

Tab.~\ref{tab:blender} reports the pose error and novel-view synthesis quality of the NeRF-Synthetic dataset. Our method achieves the smallest pose error in 5 out of 8 scenes and achieves the best reconstruction quality in all eight scenes, and the quantitative results are shown in Fig.~\ref{fig:model_comparison}. 

Tab.~\ref{tab:llff} reports the pose error and novel-view synthesis quality of the LLFF dataset. Our method achieves pose error on par with previous methods and produces the best average view synthesis quality. Our method also scores the best LPIPS in 7 out of 8 scenes, indicating that our method produces perceptually more natural novel-view synthesis.

Note that we achieve state-of-the-art results within only 20\% to 25\% of training iterations, while all other competing methods train for 200k iterations.

\input{CameraReady/LaTeX/Tables/ablations}

\input{CameraReady/LaTeX/Tables/ablation_BARF_GARF}
\input{CameraReady/LaTeX/Tables/ablation_Filters}
\input{CameraReady/LaTeX/Tables/ablation_synthetic_random_edge}

\input{CameraReady/LaTeX/Tables/ablation_sensitivity}
% Figure: 3D Qualitative Results

\subsection{Ablation}\label{exp:ablation_component}
\subsubsection{Component Analysis.}
In Tab.~\ref{tab:ablation}, we report the effect of each proposed component on the pose error and PSNR of the optimization results. The results are average across all real-world scenes in the LLFF dataset. In (a) (b), we show the effect of \emph{randomly scaled kernel} described in Sec.~\ref{method_Robust}. In (b)(c), we show the effectiveness of \emph{edge guided loss} (Sec.~\ref{method_Robust}). Finally, in (c)(d)(e), we show the necessity of Gaussian filtering on both 2D supervising images and 3D radiance field represented by a decomposed tensor grid, which validates the analysis in Sec.~\ref{method_1D}.

\subsubsection{Potential Baseline of TensoRF with BARF/GARF.}
One may suspect that we can solve the joint optimization problem of \emph{decomposed low-rank tensor} by simply applying the method of \cite{lin2021barf} or \cite{chng2022gaussian}, we clarify that there exists no simple way of integrating BARF (i.e., gradually activating higher-frequency components in positional encoding) into TensoRF since the MLP decoder of TensoRF does not take spatial coordinates as input (i.e., controlling spatial property in TensoRF is hard to achieve by manipulating positional encoding). Nevertheless, we make rough attempts to add a positional encoding schedule into the  MLP decoder input to simulate the setting of BARF or replace the decoder with a GARF network. We conduct experiments on four randomly chosen scenes in the LLFF dataset. The results are shown in Tab. \ref{tab:ablation_BaRF_GaRF}, which demonstrate the efficacy and pertinency of our proposed method to achieve successful training.

\subsubsection{Using Other Low-Pass Filters.}
% As we would like to have identical filtering strength along all spatial directions, we adopt the Gaussian filter in our method as it is the only kernel that is both circularly symmetric and separable (a well-known property in signal processing). Nevertheless, we follow the reviewer's suggestion to experiment with other low-pass filters. We report in Tab. \ref{tab:ablation_filters} the performance of using the box filter (i.e., a representative low-pass filter) on the LLFF Fortress scene, in which we clearly observe the benefits of using the Gaussian filter.
As we would like to have identical filtering strength along all spatial directions, we adopt the Gaussian filter in our method as it is the only kernel that is both circularly symmetric and separable (a well-known property in signal processing). Nevertheless, we experiment with other low-pass filters. We report in Tab. \ref{tab:ablation_filters} the performance of using the box filter (i.e., a representative low-pass filter) on the LLFF Fortress scene, in which we clearly observe the benefits of using the Gaussian filter.

\subsubsection{Applying \textit{Randomly Scaled Kernel Parameter.}
and \textit{Edge Guided Loss} on Synthetic Scenes.}
Although the two techniques are originally proposed to improve the robustness of complex real-world scenes, they do not harm the performance of synthetic ones and even slightly boost the pose estimation, as shown in Tab. \ref{tab:ablation_synthetic_random_edge}. 

\subsubsection{Sensitivity w.r.t. Pose Initialization.}
We adopt the Chair scenes in the Blender dataset to conduct sensitivity analysis upon pose initialization via varying variance $\sigma$ of Gaussian noise. The result is shown in Tab. \ref{tab:ablation_sensitivity}, which demonstrates that both BARF and our proposed method show certain robustness against the noisy initialization of camera poses.

\begin{figure}[!ht]
\centering
% \resizebox{\columnwidth}{!}{%
\includegraphics[width=1.\columnwidth]{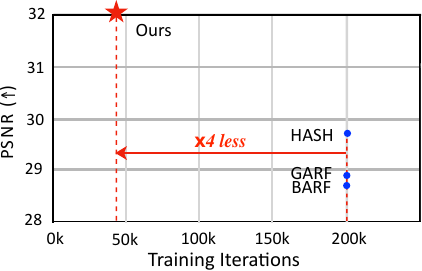}
% }
% \vskip -0.1in
\caption{\textbf{PSNR and training iterations comparison.}}
  % \vskip -0.2in
\label{fig:time}
\end{figure}

\subsection{Time Complexity} \label{exp:time}
In Fig.~\ref{fig:time}, we compare with previous methods on average PSNR and training iterations in the \emph{Synthetic NeRF} dataset. The figure shows two advantages of our method: (1) rapid convergence and (2) high-quality novel view synthesis.

% These two are closely related; in which the blurry supervision in the early stage tends to create an obstacle to detailed structure reconstruction in the latter stages, thus a long and slow joint reconstruction can hurt the quality of the final result. 
The early-stage blurry supervision can hinder detailed structure reconstruction later in the optimization, impacting the final result quality.
Our method resolves this problem by applying 3D filters with directly controllable kernel parameters, which enables smooth and rapid transition (by continuous exponential kernel schedule) of the 3D content across the spectrum domains, as opposed to previous methods that use indirect methods (e.g., learning rate in \cite{heo2023robust}, encoding magnitude in \cite{lin2021barf}) to influence learned 3D scene spectral property. 
Furthermore, our method is carefully designed to use a single voxel grid, which is trained only once in the coarse-to-fine schedule controlled by our proposed efficient component-wised convolution algorithm, thus leading to faster convergence; in comparison, \cite{heo2023robust}, which also uses voxel-based representation, requires sequential curriculum learning upon multiple voxel grids of different resolutions, resulting in four times more training iterations than ours.

% About overall speed comparisons with \cite{heo2023robust}
\begin{comment}
Compared with our results on \emph{ decomposed low-rank tensor}, \cite{heo2023robust} can achieve faster overall training speed than our method (about 25 mins vs 80 mins on synthetic scenes). However, as \cite{heo2023robust} is implemented with custom cuda-accelerated library \textit{tiny-cuda-nn} \cite{tiny-cuda-nn} (while our method is implemented purely in \text{pytorch} \cite{NEURIPS2019_9015_pytorch}), and that our method converges within 40k training iterations (compared to \cite{heo2023robust} who needs 200k training iterations), we believe that our method has the potential to achieve similar speed with optimized custom implementation.
\end{comment}

\begin{comment}
    \begin{figure}[h!]
    % \vskip 0.2in
    \begin{center}
    \centerline{\includegraphics[width=0.9\columnwidth]{CameraReady/LaTeX/Figures/TimeCompare.pdf}}
    % \vskip -0.15em
    \caption{\textbf{PSNR and training iterations comparison.}~\yulunliu{Fig 5 and Fig 6 side-by-side.}}
    \label{fig:time}
    \end{center}
    % \vskip -0.2in
\end{figure}
\end{comment}

%% file: CameraReady/LaTeX/Tables/synthetic.tex
\setlength{\tabcolsep}{3pt}
\begin{table*}[tbh!]
  \centering
  % \small
  \footnotesize
  % \vspace{-0.7em}
  
  % \vskip 0.13in
  % \begin{adjustbox}{width=1\textwidth} 
  % \setlength{\tabcolsep}{0pt}
  \begin{tabular}{ l  c c c c c c c c   c c c c c c c c c c c}
    \toprule
    
    \multirow{4}{*}{Scene} 
    & \multicolumn{8}{c}{Camera Pose Registration} 
    & \multicolumn{8}{c}{View Synthesis Quality} \\ 
    
    \cmidrule(r){2-9}  
    \cmidrule(r){10-17} 
    & \multicolumn{4}{c}{Rotation $({}^{\circ})$ $\downarrow$}
    & \multicolumn{4}{c}{Translation $\downarrow$}
    & \multicolumn{4}{c}{PSNR $\uparrow$}
    & \multicolumn{4}{c}{SSIM $\uparrow$} \\
    \cmidrule(r){2-5}
    \cmidrule(r){6-9}
    \cmidrule(r){10-13}
    \cmidrule(r){14-17}
    & GARF
    & BARF
    & HASH
    & Ours
    & GARF
    & BARF
    & HASH
    & Ours
    & GARF
    & BARF
    & HASH
    & Ours
    & GARF
    & BARF
    & HASH
    & Ours
    \\
    \midrule 
    
    Chair 
    & 0.113
    & 0.096
    & \textbf{0.085}
    % & 0.8743
    & 0.874
    
    & 0.549
    & 0.428
    & \textbf{0.365}
    & 3.501
    
    & 31.32
    & 31.16
    & 31.95
    % & \textbf{35.227}
    & \textbf{35.22}
    
    & 0.959
    & 0.954
    & 0.962
    & \textbf{0.984}
    \\
    
    Drum 
    & 0.052
    & 0.043
    & 0.041
    & \textbf{0.037}
    
    & 0.232
    & 0.225
    & 0.214
    & \textbf{0.118}
    
    & 24.15
    & 23.91
    & 24.16
    & \textbf{25.78}
    
    & 0.909
    & 0.900
    & 0.912
    & \textbf{0.934}
    \\
    
    Ficus 
    & 0.081
    & 0.085
    &  0.079
    & \textbf{0.050}
    
    & 0.461
    & {0.474}
    & 0.479
    & \textbf{0.173}
    
    & 26.29
    & 26.26
    & 28.31
    & \textbf{31.37}
    
    & 0.935
    & 0.934
    & 0.943
    & \textbf{0.978}
    \\
    
    Hotdog 
    & 0.235
    & 0.248
    & 0.229
    & \textbf{0.105}
    
    & 1.123
    & 1.308
    & 1.123
    & \textbf{0.499}
    
    & 34.69
    & 34.54
    & 35.41
    & \textbf{37.18}
    
    & 0.972
    & 0.970
    & 0.981
    & \textbf{0.982}
    \\
    
    Lego 
    & 0.101 
    & 0.082
    & 0.071
    & \textbf{0.049}
    
    & 0.299
    & 0.291
    & 0.272
    & \textbf{0.100}
    
    & 29.29
    & 28.33
    & 31.65
    & \textbf{34.23}
    
    & 0.925
    & 0.927
    & 0.973
    & \textbf{0.981}
    \\
    
    Materials 
    & \textbf{0.842}
    & {0.844}
    & 0.852
    & 0.854
    
    & \textbf{2.688}
    & {2.692}
    & 2.743
    & 2.690
    
    & 27.91
    & {27.84}
    & 27.14
    & \textbf{29.04}
    
    & 0.941
    & 0.936 
    & 0.911
    & \textbf{0.951}
    \\
    
    Mic 
    & 0.070
    & 0.071
    & \textbf{0.068}
    & 1.177
    
    & 0.293
    & 0.301
    & \textbf{0.287}
    & 5.000
    
    & 31.39
    & 31.18
    & 32.33
    & \textbf{32.50}
    
    & 0.971
    & 0.969
    & 0.975
    & \textbf{0.976}
    \\
    
    Ship 
    & 0.073
    & {0.075}
    & 0.079
    & \textbf{0.058}
    
    & 0.310
    & 0.326
    & 0.287
    & \textbf{0.167}
    
    & 27.64
    & 27.50
    & 27.92
    & \textbf{31.98}
    
    & 0.862
    & 0.849
    & 0.879
    & \textbf{0.903}
    \\
    
    \midrule
    Mean 
    & 0.195
    & 0.193
    & \textbf{0.189}
    & 0.400
    
    & 0.744
    & 0.756
    & \textbf{0.722}
    & 1.533
    
    & 28.96
    & 28.84
    & 29.86
    & \textbf{32.07}
    
    & 0.935
    & 0.930
    & 0.943
    & \textbf{0.961}
    \\
    
    \bottomrule
  \end{tabular}
  % \end{adjustbox}
  % \vskip -0.05in
     \caption{
  \textbf{Quantitative results on the NeRF-Synthetic dataset.} Our method achieves the best average novel-view synthesis quality and the best pose error in 5 out of 8 scenes. Notice that our method converges within 40k iterations, while all previous methods train for 200k iterations.
  }
  \label{tab:blender}
  % \vskip -0.1in
\end{table*}

%% file: CameraReady/LaTeX/Tables/llff.tex
\setlength{\tabcolsep}{3pt}
\begin{table*}[tbh!]
  \centering
  % \small
  \footnotesize
  % \vskip 0.13in
  % \begin{adjustbox}{width=1\textwidth} 
  \begin{tabular}{ l  c c c c  c c c c   c c c c  c c c c }
    \toprule
    
    \multirow{4}{*}{Scene} 
    & \multicolumn{8}{c}{Camera Pose Registration} 
    & \multicolumn{8}{c}{View Synthesis Quality} \\ 
    
    \cmidrule(r){2-9}  
    \cmidrule(r){10-17} 
    & \multicolumn{4}{c}{Rotation $({}^{\circ})$ $\downarrow$}
    & \multicolumn{4}{c}{Translation $\downarrow$}
    & \multicolumn{4}{c}{PSNR $\uparrow$}
    & \multicolumn{4}{c}{SSIM $\uparrow$}\\
    \cmidrule(r){2-5}
    \cmidrule(r){6-9}
    \cmidrule(r){10-13}
    \cmidrule(r){14-17}
    & GARF
    & BARF
    & HASH
    & Ours
    
    & GARF
    & BARF
    & HASH
    & Ours
    
    & GARF
    & BARF
    & HASH
    & Ours
    
    & GARF
    & BARF
    & HASH
    & Ours
    \\
    \midrule 
    
    Fern 
    & 0.470
    & 0.191
    & \textbf{0.110}
    & 0.472
    
    & 0.250
    & \textbf{0.102}
    & \textbf{0.102}
    & 0.199
    
    & 24.51
    & 23.79
    & 24.62
    & \textbf{26.17}
    
    & 0.740
    & 0.710
    & 0.743
    & \textbf{0.842}
    \\
    
    Flower 
    & 0.460
    & \textbf{0.251} % 0.47?
    & 0.301
    & 1.375
    
    & 0.220
    & 0.224
    & \textbf{0.211}
    & 0.389
    
    & \textbf{26.40}
    & 23.37
    & 25.19
    & 25.62
    
    & 0.790
    & 0.698
    & 0.744
    & \textbf{0.810}
    \\
    
    Fortress 
    & \textbf{0.030}
    & 0.479 % 0.17?
    & 0.211
    & 0.449
    
    & 0.270
    & 0.364
    & \textbf{0.241}
    & 0.419
    
    & 29.09
    & 29.08
    & \textbf{30.14}
    & 29.68
    
    & 0.820
    & 0.823
    & \textbf{0.901}
    & 0.882
    \\
    
    Horns 
    & \textbf{0.030}
    & 0.304 % 3.50? 
    & 0.049
    & 0.386
    
    & 0.210 
    & 0.222 % 1.32?
    & \textbf{0.209}
    & 0.251
    
    & 22.54
    & 22.78
    & \textbf{22.97}
    & 22.84
    
    & 0.690
    & 0.727
    & 0.736
    & \textbf{0.819}
    \\
    
    Leaves 
    & \textbf{0.130}
    & 1.272
    & 0.840
    & 1.990
    
    & 0.230
    & 0.249
    & \textbf{0.228}
    & 0.397
    
    & 19.72
    & 18.78
    & 19.45
    & \textbf{21.24}
    
    & 0.610
    & 0.537
    & 0.607
    & \textbf{0.753}
    \\
    
    Orchids 
    & 0.430
    & 0.627
    & 0.399
    & \textbf{0.279}
    
    & 0.410
    & 0.404
    & 0.386
    & \textbf{0.340}
    
    & 19.37
    & 19.45
    & 20.02
    & \textbf{20.57}
    
    & 0.570
    & 0.574
    & 0.610
    & \textbf{0.698}
    \\
    
    Room 
    & 0.270
    & 0.320
    & 0.271
    & \textbf{0.188}
    
    & 0.200
    & 0.270
    & 0.213
    & 0.191
    
    & 31.90
    & 31.95
    & \textbf{32.73}
    & 31.87
    
    & 0.940
    & {0.949}
    & \textbf{0.968}
    & 0.936
    \\
    
    T-Rex 
    & \textbf{0.420}
    & 1.138 % 0.66?
    & 0.894
    & 0.523
    
    & \textbf{0.360}
    & 0.720 % 0.36?
    & 0.474
    & 0.416
    
    & 22.86
    & 22.55
    & 23.19
    & \textbf{24.19}
    
    & 0.800
    & 0.767
    & 0.866
    & \textbf{0.878}
    \\
    
    \midrule
    Mean 
    & \textbf{0.280}
    & 0.573
    & 0.384
    & 0.709
    
    & 0.269
    & 0.331
    & \textbf{0.258}
    & 0.325
    
    & 24.55
    & 23.97
    & 24.79
    & \textbf{25.27}
    
    & 0.745
    & 0.723
    & 0.772
    & \textbf{0.827}
    \\
    
    \bottomrule
  \end{tabular}
  % \end{adjustbox}
  % \vskip -0.05in
  \caption{
      \textbf{Quantitative results on the LLFF dataset.} 
      % ~\yulunliu{I feel we don't need to mention this?} ~\boyu{agree. I also fix the main text description.}
      Our method achieves the best average novel-view synthesis quality and best LPIPS in 7 out of 8 scenes. Our method converges within 50k iterations, while all previous methods train for 200k iterations.
  }
  \label{tab:llff}
\end{table*}

%% file: CameraReady/LaTeX/Tables/ablations.tex
\setlength{\tabcolsep}{1pt}
\begin{table}[t!]
  \centering
  % \small
  \footnotesize
  %\vspace{-0.7em}  % shrink verticle width

  % \vskip 0.15in
  % \resizebox{\columnwidth}{!}{%
  \begin{tabular}{ c  c c c c | c c c  }
    \toprule
    % \multirow{2}{*}{} 
    % & \multicolumn{4}{c}{\textbf{Component Ablation}} 
    % & \multicolumn{3}{c}{\textbf{Evaluation Metric}} 
    % \\
    % \cmidrule{2-5} 
    % \cmidrule{6-8}
    & $\ $ 3D $\ $
    & $\ $ 2D $\ $
    & $\ $ Random $\ $
    & $\ $ Edge $\ $
    % \hspace{1pt} 
    & {Rot.} 
    & {Trans.} 
    & {PSNR} 
    \\  
    & $\ $ Gauss. $\ $
    & $\ $ Gauss. $\ $
    & $\ $ Kernel $\ $
    & $\ $ Guided $\ $
    % \hspace{1pt} 
    & {$\downarrow$} 
    & {$\downarrow$} 
    & {$\uparrow$} 
    \\  
    \midrule
     
    (a)
    & \checkmark & \checkmark & \checkmark &\checkmark 
    & \textbf{0.72}
    & \textbf{0.33}
    & \textbf{25.36}
    \\ 
    
    (b)
    &  \checkmark & \checkmark &  & \checkmark
    & 1.00
    % & 0.371
    & 0.37
    & 25.25
    \\
    
    (c)
    &  \checkmark & \checkmark  &  &
    & 1.91
    % & 0.939
    & 0.93
    & 25.12
    \\
     
    (d)
    & \checkmark &   &  & 
    & 33.00
    & 12.7
    & 20.10
    \\

    (e) 
    &  &\checkmark  &  &
    & 26.25
    & 8.9
    & 19.73
    \\

    (d) 
    &  &   &  &
    & 23.29
    & 9.4
    & 23.97
    \\

    \bottomrule
  \end{tabular}
  % }
  % \vskip -0.05in
    \caption{\textbf{Ablation study of the components of the proposed method on the real-world LLFF dataset.} 
    % Experiments are conducted on the real-world LLFF dataset.
    % Four components are the \emph{3D Gaussian kernel} in Section \ref{method_3D}, \emph{2D Gaussian kernel} for smoothed supervision, \emph{randomized kernel parameter}, and \emph{edge-guided rendering loss} in  Section\ref{method_Robust}. 
  }
  % \vskip -0.1in
  \label{tab:ablation}
  % \vspace{1pt}
\end{table}

%% file: CameraReady/LaTeX/Tables/ablation_BARF_GARF.tex
\setlength{\tabcolsep}{2pt}
\begin{table}[t!]
  \centering
  % \small
  \footnotesize
  %\vspace{-0.7em}  % shrink verticle width
    % \resizebox{0.99\columnwidth}{!}{%
        \begin{tabular}{l|ccccc}
        \toprule
         & Rot. $\downarrow$ & Trans. $\downarrow$  & PSNR $\uparrow$ & SSIM $\uparrow$ & LPIPS $\downarrow$\\ 
        \midrule
        TensoRF + BARF & 45.47 & 0.17 & 20.71 & 0.630 & 0.314\\
        TensoRF + GARF & 73.92 & 0.29 & 10.47 & 0.287& 0.679\\
        Ours & 0.43 & 0.003 & 26.92 & 0.872 & 0.104\\
        \bottomrule
        \end{tabular}
    % }

  % \vskip 0.15in
  % \vskip -0.05in
    \caption{\textbf{Ablation on Directly Applying BARF and GARF on TensoRF (Potential Baseline)} 
    % Experiments are conducted on the real-world LLFF dataset.
    % Four components are the \emph{3D Gaussian kernel} in Section \ref{method_3D}, \emph{2D Gaussian kernel} for smoothed supervision, \emph{randomized kernel parameter}, and \emph{edge-guided rendering loss} in  Section\ref{method_Robust}. 
  }
  % \vskip -0.1in
  \label{tab:ablation_BaRF_GaRF}
  % \vspace{1pt}
\end{table}

%% file: CameraReady/LaTeX/Tables/ablation_Filters.tex
\setlength{\tabcolsep}{4pt}
\begin{table}[t!]
  \centering
  % \small
  \footnotesize
  %\vspace{-0.7em}  % shrink verticle width
    % \resizebox{0.99\columnwidth}{!}{%
        \begin{tabular}{l|ccccc}
        \toprule
        Filter & Rot. $\downarrow$ & Trans. $\downarrow$ & PSNR $\uparrow$ & SSIM $\uparrow$& LPIPS $\downarrow$\\
        \midrule
        Box filter & 9.98 & 0.06 & 20.18 & 0.387 & 0.165 \\
        Gaussian filter & 0.46 & 0.004 & 29.49 & 0.874 & 0.063\\
        \bottomrule
        \end{tabular}
    % }
  % \vskip 0.15in
  % \vskip -0.05in
    \caption{\textbf{Ablation On Low-Pass Filters.} 
    % Experiments are conducted on the real-world LLFF dataset.
    % Four components are the \emph{3D Gaussian kernel} in Section \ref{method_3D}, \emph{2D Gaussian kernel} for smoothed supervision, \emph{randomized kernel parameter}, and \emph{edge-guided rendering loss} in  Section\ref{method_Robust}. 
  }
  % \vskip -0.1in
  \label{tab:ablation_filters}
  % \vspace{1pt}
\end{table}

%% file: CameraReady/LaTeX/Tables/ablation_synthetic_random_edge.tex
\setlength{\tabcolsep}{2pt}
\begin{table}[t!]
  \centering
  % \small
  \footnotesize
    % \resizebox{0.99\columnwidth}{!}{%
        \begin{tabular}{l|ccc}
        \toprule
        % Setting (4 randomly chosen scenes from LLFF) & Rot. $\downarrow$ & Trans. $\downarrow$ & PSNR $\uparrow$\\
        Setting & Rot. $\downarrow$ & Trans. $\downarrow$ & PSNR $\uparrow$\\
        \midrule
        % w/ randomly scaled kernel \& edge guided loss & 0.06 & 0.002 & 34.34 \\
        % w/o randomly scaled kernel \& edge guided loss  & 0.28 & 0.010 & 34.43 \\
        w/ random kernels \& edge guided loss & 0.06 & 0.002 & 34.34 \\
        w/o random kernels \& edge guided loss  & 0.28 & 0.010 & 34.43 \\
        \bottomrule
        \end{tabular}
    % }
  % \vskip 0.15in
  % \vskip -0.05in
    \caption{\textbf{Ablation on Applying \textit{Randomly Scaled Kernel Parameter} and \textit{Edge Guided Loss} in Synthetic Scenes} 
    % Experiments are conducted on the real-world LLFF dataset.
    % Four components are the \emph{3D Gaussian kernel} in Section \ref{method_3D}, \emph{2D Gaussian kernel} for smoothed supervision, \emph{randomized kernel parameter}, and \emph{edge-guided rendering loss} in  Section\ref{method_Robust}. 
  }
  % \vskip -0.1in
  \label{tab:ablation_synthetic_random_edge}
  % \vspace{1pt}
\end{table}

%% file: CameraReady/LaTeX/Tables/ablation_sensitivity.tex
\setlength{\tabcolsep}{6pt}
\begin{table}[t!]
  \centering
  % \small
  \footnotesize
  % \vspace{-3pt}
    % \resizebox{0.9\columnwidth}{!}{%
        \begin{tabular}{lr|cccc}
        \hline
        \toprule
            \multicolumn{2}{r|}{$\sigma$} & $0.125$ & $0.15$ & $0.175$ & $0.2$ \\
            \midrule
            % $\sigma$ & & $0.125$ & $0.15$ & $0.175$ & $0.2$ & $0.25$ \\
            \multirow{2}{*}{BARF}& Rotation $\downarrow$ & 0.094 & 0.068 & 0.100 & 0.108 \\
            &Translation $\downarrow$ & 0.004 & 0.004 & 0.005 & 0.005 \\
            \midrule
            \multirow{2}{*}{Ours}& Rotation $\downarrow$ & 0.07 & 0.062 & 0.072 & 0.066 \\

            &Translation $\downarrow$ & 0.003 & 0.003 & 0.003 & 0.002 \\
        \bottomrule
        \end{tabular}
    % }
  % \vskip 0.15in
  % \vskip -0.05in
    \caption{\textbf{Ablation: Sensitivity Analysis On Gaussian Noise in Blender Chair.} 
    % Experiments are conducted on the real-world LLFF dataset.
    % Four components are the \emph{3D Gaussian kernel} in Section \ref{method_3D}, \emph{2D Gaussian kernel} for smoothed supervision, \emph{randomized kernel parameter}, and \emph{edge-guided rendering loss} in  Section\ref{method_Robust}. 
  }
  % \vskip -0.1in
  \label{tab:ablation_sensitivity}
  % \vspace{1pt}
\end{table}

%% file: CameraReady/LaTeX/conclusion.tex
\section{Conclusion}
Our contributions is three fold:
1) \emph{Theoretically}, we provide insights into the impact of 3D scene properties on the convergence of joint optimization beyond the coarse-to-fine heuristic discussed in prior research (e.g., BARF, Heo \etal~2023), thus offering a filtering strategy for improving the joint optimization of camera pose and 3D radiance field.
2) \emph{Algorithmically}, we introduce (and prove the equivalence of) an effective method for applying the pilot study's filtering strategy on the decomposed low-rank tensor, notice that the proposed \textit{separable component-wise convolution} is more efficient than the traditionally well-known trick of \textit{separable convolution kernel} as we additionally utilize the separability of the input signal. Furthermore, we also propose other techniques such as \textit{randomly-scaled kernel parameter}, \textit{blurred 2D supervision}, and \textit{edge-guided loss mask} to help our proposed method better perform in complex real-world scenes.
3)
Comprehensive evaluations demonstrate our proposed framework’s state-of-the-art performance and rapid convergence without known poses.